\documentclass{article} 
\usepackage{iclr2019_conference,times}

\usepackage{amsmath,amsfonts,bm}

\def\eqref#1{equation~\ref{#1}}

\def\1{\bm{1}}

\def\ri{{\textnormal{i}}}

\DeclareMathAlphabet{\mathsfit}{\encodingdefault}{\sfdefault}{m}{sl}
\SetMathAlphabet{\mathsfit}{bold}{\encodingdefault}{\sfdefault}{bx}{n}

\newcommand{\R}{\mathbb{R}}

\usepackage{hyperref}
\usepackage{url}
\usepackage{comment}
\usepackage{amsmath,amssymb,amsfonts}
\usepackage{graphicx}
\usepackage{wrapfig}
\usepackage{paralist}
\usepackage{subfig}
\usepackage{comment}
\usepackage{algorithm}
\usepackage{algpseudocode}
\usepackage{booktabs}

\newcommand{\note}[1]{{\it \color{red} [#1]}}

\newcommand{\sched}{SchedNet}
\newcommand{\myparagraph}[1]{\vspace{-0.25cm} \paragraph{#1}}

\title{Learning to Schedule Communication \\ in Multi-agent Reinforcement Learning}

\author{Daewoo Kim, Sangwoo Moon, David Hostallero, Wan Ju Kang, Taeyoung Lee, \\ {\bf Kyunghwan Son \& Yung Yi} \\
School of Electrical Engineering, KAIST\\
Daejeon, South Korea\\
\texttt{\small \{dwkim, swmoon, ddhostallero, wjkang, taeyoung.lee, khson\}@lanada.kaist.}\\ \texttt{\small ac.kr, yiyung@kaist.edu} \\
}

\iclrfinalcopy 
\begin{document}

\newcommand{\real}{{\mathbb R}}
\newcommand{\integer}{{\mathbb Z}}



\newcommand{\svq}{\ensuremath{P_s}}
\newcommand{\svc}{\ensuremath{P_0}}
\newcommand{\svs}{\ensuremath{Q_s}}
\newcommand{\svd}{\ensuremath{D}}
\newcommand{\svdd}{\ensuremath{4 \svq \svs - \svc^2}}

\newcommand{\sqd}{\ensuremath{D'}}
\newcommand{\sqdd}{\ensuremath{4 \svq \svs - \eta^2\svc^2}}

\newcommand{\siv}{\ensuremath{{x}}}
\newcommand{\sic}{\ensuremath{{A}}}

\newcommand{\bx}{\bm{x}}
\newcommand{\bc}{\bm{c}}
\newcommand{\bbx}{\ensuremath{{\bm x}}}
\newcommand{\bbc}{\ensuremath{{\bm c}}}
\newcommand{\bu}{\ensuremath{{\bm u}}}

\newcommand{\bz}{\ensuremath{{\bm z}}}
\newcommand{\ba}{\ensuremath{{\bm a}}}
\newcommand{\be}{\ensuremath{{\bm e}}}
\newcommand{\br}{\ensuremath{{\bm r}}}
\newcommand{\by}{\ensuremath{{\bm y}}}
\newcommand{\bby}{\ensuremath{{\bm y}}}
\newcommand{\bbq}{\ensuremath{{\bm q}}}

\newcommand{\bp}{\ensuremath{{\bm p}}}
\newcommand{\bh}{\ensuremath{{\bm h}}}
\newcommand{\bP}{\bm{P}}
\newcommand{\bA}{\ensuremath{{\bm A}}}

\newcommand{\bF}{\bm{F}}
\newcommand{\bN}{\ensuremath{{\bm N}}}
\newcommand{\bbf}{\ensuremath{{\bm f}}}
\newcommand{\bG}{\ensuremath{{\bm G}}}
\newcommand{\bQ}{\ensuremath{{\bm Q}}}
\newcommand{\bH}{\ensuremath{{\bm H}}}
\newcommand{\bW}{\ensuremath{{\bm W}}}
\newcommand{\bJ}{\ensuremath{{\bm J}}}
\newcommand{\bg}{\ensuremath{{\bm g}}}
\newcommand{\bs}{\ensuremath{{\bm s}}}

\newcommand{\bbv}{\ensuremath{{\bm v}}}
\newcommand{\bbz}{\ensuremath{{\bm z}}}
\newcommand{\bbu}{\ensuremath{{\bm u}}}
\newcommand{\bbm}{\ensuremath{{\bm m}}}
\newcommand{\bbU}{\ensuremath{{\bm U}}}
\newcommand{\bw}{\bm{w}}
\newcommand{\bbr}{\ensuremath{{\bm r}}}
\newcommand{\bbalpha}{\ensuremath{{\bm \alpha}}}
\newcommand{\brho}{\ensuremath{{\bm \rho}}}



\newcommand{\bR}{\bm{R}}

\newcommand{\cR}{\ensuremath{\mathcal R}}
\newcommand{\set}[1]{\ensuremath{\mathcal #1}}
\renewcommand{\vec}[1]{\bm{#1}}
\newcommand{\mat}[1]{\bm{#1}}

\newcommand{\node}[1]{\ensuremath{{\tt #1}}}
\newcommand{\nn}[1]{\node{#1}}

\newcommand{\singleq}[1]{\text{`}#1\text{'}}

\newcommand{\mytitle}[1]{\medskip \noindent{\em #1} \smallskip}
\newcommand{\mytitlehead}[1]{\noindent{\em #1} \smallskip}

\newcommand{\separator}{
  \begin{center}
    \rule{\columnwidth}{0.3mm}
  \end{center}
}
\newenvironment{separation}{ \vspace{-0.3cm}  \separator  \vspace{-0.2cm}}
{  \vspace{-0.4cm}  \separator  \vspace{-0.1cm}}

\def\un{\underline}
\def\ov{\overline}
\def\Bl{\Bigl}
\def\Br{\Bigr}
\def\lf{\left}
\def\ri{\right}
\def\st{\star}

\def\eg{{\it e.g.}}
\def\ie{{\it i.e.}}

\newtheorem{theorem}{Theorem}[section]
\newtheorem{corollary}{Corollary}[section]
\newtheorem{property}{Property}[section]
\newtheorem{proposition}{Proposition}[section]
\newtheorem{lemma}{Lemma}[section]
\newtheorem{definition}{Definition}[section]
\newtheorem{condition}{Condition}[section]
\newtheorem{assumption}{Assumption}[section]
\newtheorem{example}{Example}[section]
\newtheorem{remark}{Remark}[section]
\newtheorem{problem}{Problem}[section]

\newcommand{\bprob}[1]{\mathbb{P}\Bl[ #1 \Br]}
\newcommand{\prob}[1]{\mathbb{P}[ #1 ]}
\newcommand{\expect}[1]{\mathbb{E}[ #1 ]}
\newcommand{\bexpect}[1]{\mathbb{E}\Bl[ #1 \Br]}
\newcommand{\bbexpect}[1]{\mathbb{E}\lf[ #1 \ri]}
\newcommand{\vari}[1]{\text{VAR}[ #1 ]}
\newcommand{\bvari}[1]{\text{VAR}\Bl[ #1 \Br]}

\newcommand{\exming}{\ensuremath{\text{ExMin}_g} }
\newcommand{\imp}{\Longrightarrow}
\newcommand{\beq}{\begin{eqnarray*}}
\newcommand{\eeq}{\end{eqnarray*}}
\newcommand{\beqn}{\begin{eqnarray}}
\newcommand{\eeqn}{\end{eqnarray}}
\newcommand{\bemn}{\begin{multiline}}
\newcommand{\eemn}{\end{multiline}}

\newcommand{\sqeq}{\addtolength{\thinmuskip}{-4mu}
\addtolength{\medmuskip}{-4mu}\addtolength{\thickmuskip}{-4mu}}

\newcommand{\unsqeq}{\addtolength{\thinmuskip}{+4mu}
\addtolength{\medmuskip}{+4mu}\addtolength{\thickmuskip}{+4mu}}


\newcommand{\grad}[1]{\nabla #1}







\def\A{\mathcal A}
\def\oA{\overline{\mathcal A}}
\def\S{\mathcal S}
\def\D{\mathcal D}
\def\eff{{\rm Eff}}

\def\cU{{\cal U}}
\def\cM{{\cal M}}
\def\cV{{\cal V}}
\def\cA{{\cal A}}
\def\cX{{\cal X}}
\def\cN{{\cal N}}
\def\cJ{{\cal J}}
\def\cK{{\cal K}}
\def\cL{{\cal L}}
\def\cI{{\cal I}}
\def\cY{{\cal Y}}
\def\cZ{{\cal Z}}
\def\cC{{\cal C}}
\def\cR{{\cal R}}
\def\id{{\rm Id}}
\def\st{{\rm st}}
\def\cF{{\cal F}}
\def\cG{{\cal G}}
\def\N{\mathbb{N}}
\def\R{\mathbb{R}}
\def\cB{{\cal B}}
\def\cP{{\cal P}}
\def\cS{{\cal S}}
\def\cT{{\cal T}}
\def\cO{{\cal O}}
\def\ind{{\bf 1}}

\def\bmg{{\bm{\gamma}}}
\def\bmr{{\bm{\rho}}}
\def\bmq{{\bm{q}}}
\def\bmt{{\bm{\tau}}}
\def\bmn{{\bm{n}}}
\def\bmcapn{{\bm{N}}}
\def\bmrho{{\bm{\rho}}}

\def\igam{\underline{\gamma}(\lambda)}
\def\sgam{\overline{\gamma}(\lambda)}

\def\PP{{\mathrm P}}
\def\EE{{\mathrm E}}
\def\iskip{{\vskip -0.4cm}}
\def\siskip{{\vskip -0.2cm}}

\def\bp{\noindent{\it Proof.}\ }
\def\ep{\hfill $\Box$}

\def\sout#1{{\leavevmode\setbox0\hbox{#1}%
    \rlap{\vrule width\wd0 height.8ex depth-.6ex}\box0 }}

\def\soutr#1{{\leavevmode\setbox0\hbox{#1}%
    \rlap{\textcolor{red}{\vrule width\wd0 height.8ex depth-.6ex}}\box0 }}

\maketitle

\begin{abstract}
  Many real-world reinforcement learning tasks require multiple agents
  to make sequential decisions under the agents' interaction, where
  well-coordinated actions among the agents are crucial to achieve the
  target goal better at these tasks. One way to accelerate the
  coordination effect is to enable multiple agents to communicate with
  each other in a distributed manner and behave as a group.
  In this paper, we study a practical scenario when {\em (i)} the
  communication bandwidth is limited and {\em (ii)} the agents share the
  communication medium so that only a restricted number of agents are
  able to simultaneously use the medium, as in the state-of-the-art wireless networking standards.  
  This calls for a certain
  form of {\em communication scheduling.}  In that regard, we propose a
  multi-agent deep reinforcement learning framework, called {\sched},
  in which agents learn how to schedule themselves, how to encode the
  messages, and how to select actions based on received
  messages. {\sched} is capable of deciding which agents should be
  entitled to broadcasting their (encoded) messages, by learning the
  importance of each agent's partially observed information.  We
  evaluate {\sched} against multiple baselines under two different
  applications, namely, cooperative communication and navigation, and
  predator-prey. Our experiments show a non-negligible performance gap
  between {\sched} and other mechanisms such as the ones without
  communication and with vanilla scheduling methods, {\em e.g.}, round
  robin, ranging from 32\% to 43\%.
\end{abstract}

\section{Introduction}
\label{sec:intro}
Reinforcement Learning (RL) has garnered renewed interest in recent
years.  Playing the game of Go \citep{dqn}, robotics control
\citep{robot,ddpg}, and adaptive video streaming \citep{pensieve}
constitute just a few of the vast range of RL applications. Combined
with developments in deep learning, deep reinforcement learning (Deep
RL) has emerged as an accelerator in related fields. From the
well-known success in single-agent deep reinforcement learning, such
as \citet{dqn}, we now witness growing interest in its multi-agent
extension, the multi-agent reinforcement learning (MARL), exemplified
in \citet{gupta2017cooperative,lowe2017multi,coma,omidshafiei2017deep,
  dial, commnet, mordatch2017grounded,havrylov2017,lenient,bicnet,
  lola, tampuu2017multiagent, leibo2017multi,foerster2017stabilising}.
In the MARL problem commonly addressed in these works, multiple agents
interact in a single environment repeatedly and improve their policy
iteratively by learning from observations to achieve a common goal.
Of particular interest is the distinction between two lines of
research: one fostering the direct communication among agents
themselves, as in \citet{dial,commnet} and the other coordinating their
cooperative behavior without direct communication, as in
\citet{foerster2017stabilising,lenient,leibo2017multi}.

In this work, we concern ourselves with the former. We consider MARL
scenarios wherein the task at hand is of a cooperative nature and
agents are situated in a partially observable environment, but each
endowed with different observation power.  We formulate this scenario
into a multi-agent sequential decision-making problem, such that all
agents share the goal of maximizing the same discounted sum of
rewards. For the agents to directly communicate with each other and
behave as a coordinated group rather than merely coexisting
individuals, they must carefully determine the information they
exchange under a practical bandwidth-limited environment and/or in the case
of high-communication cost. To coordinate this exchange of messages,
we adopt the centralized training and distributed execution paradigm
popularized in recent works, {\em e.g.}, \citet{coma, lowe2017multi, vdn,
  qmix, gupta2017cooperative}.

In addition to bandwidth-related constraints, we take the issues of
sharing the communication medium into consideration, especially when
agents communicate over wireless channels. The state-of-the-art standards on wireless communication such as Wi-Fi and LTE specify the way of scheduling users as one of the basic functions. However, as elaborated in
Related work, MARL problems involving scheduling of only a restricted set
of agents have not yet been extensively studied. The key challenges in
this problem are: {\em (i)} that limited bandwidth implies that agents must
exchange succinct information: something concise and yet meaningful
and {\em (ii)} that the shared medium means that potential contenders must
be appropriately arbitrated for proper collision avoidance,
necessitating a certain form of communication scheduling, popularly 
referred to as MAC (Medium Access Control) in the area of wireless
communication.  While stressing the coupled nature of the
encoding/decoding and the scheduling issue, we zero in on the said
communication channel-based concerns and construct our neural network
accordingly.

\myparagraph{Contributions} In this paper, we propose a new deep
multi-agent reinforcement learning architecture, called {\sched}, with
the rationale of centralized training and distributed execution in
order to achieve a common goal better via decentralized cooperation.
During distributed execution, agents are allowed to communicate over
wireless channels where messages are broadcast to all agents in each
agent's communication range. This broadcasting feature of wireless
communication necessitates a Medium Access Control (MAC) protocol to
arbitrate contending communicators in a shared medium. CSMA (Collision
Sense Multiple Access)  in Wi-Fi is one such
MAC protocol. While prior work on MARL to date considers only the
limited bandwidth constraint, we additionally address the shared
medium contention issue in what we believe is the first work of its
kind: which nodes are granted access to the shared medium.
Intuitively, nodes with more important observations should be chosen,
for which we adopt a simple yet powerful mechanism called weight-based
scheduler (WSA), designed to reconcile simplicity in training with
integrity of reflecting real-world MAC protocols in use ({\em e.g.},
802.11 Wi-Fi).  We evaluate {\sched} for two applications: cooperative
communication and navigation and predator/prey and demonstrate that
{\sched} outperforms other baseline mechanisms such as the one without
any communication or with a simple scheduling mechanism such as round
robin.  We comment that {\sched} is not intended for competing with
other algorithms for cooperative multi-agent tasks without considering
scheduling, but a complementary one. We believe that adding our idea
of agent scheduling makes those algorithms much more practical and
valuable.

\myparagraph{Related work} We now discuss the body of relevant
literature. \citet{busoniu2008} and \citet{iql} have studied MARL with
decentralized execution extensively. However, these are based on
tabular methods so that they are restricted to simple environments.
Combined with developments in deep learning, deep MARL algorithms have
emerged \citep{tampuu2017multiagent, coma, lowe2017multi}.
\citet{tampuu2017multiagent} uses a combination of DQN with
independent Q-learning. This independent learning does not perform
well because each agent considers the others as a part of environment
and ignores them. \citet{coma, lowe2017multi, gupta2017cooperative,
  vdn}, and \citet{foerster2017stabilising} adopt the framework of centralized training with
decentralized execution, empowering the agent to learn cooperative
behavior considering other agents' policies without any communication
in distributed execution.

It is widely accepted that communication can further enhance the
collective intelligence of learning agents in their attempt to
complete cooperative tasks. To this end, a number of papers have
previously studied the learning of communication protocols and
languages to use among multiple agents in reinforcement learning. We
explore those bearing the closest resemblance to our research.
\citet{dial,commnet,bicnet, guestrin2002}, and \citet{zhang2013} train multiple
agents to learn a communication protocol, and have shown that
communicating agents achieve better rewards at various tasks.  \citet
{mordatch2017grounded} and \citet{havrylov2017} investigate the possibility of the
artificial emergence of language. Coordinated RL by
\citet{guestrin2002} is an earlier work demonstrating the feasibility
of structured communication and the agents' selection of jointly
optimal action.

Only DIAL \citep{dial} and \citet{zhang2013} explicitly address
bandwidth-related concerns. In DIAL, the communication channel of the
training environment has a limited bandwidth, such that the agents
being trained are urged to establish more resource-efficient
communication protocols. The environment in \citet{zhang2013} also has
a limited-bandwidth channel in effect, due to the large amount of
exchanged information in running a distributed constraint optimization
algorithm. Recently, \citet{attentional_comm} proposes an attentional communication model
that allows some agents who request additional information from others to
gather observation from neighboring agents. However, they
do not explicitly consider the constraints imposed by limited communication bandwidth and/or scheduling due to communication over a shared medium.

To the best of our knowledge, there is no prior work that incorporates an
intelligent scheduling entity in order to facilitate inter-agent
communication in both a limited-bandwidth and shared medium access
scenarios. As outlined in the introduction, intelligent scheduling
among learning agents is pivotal in the orchestration of their
communication to better utilize the limited available
bandwidth as well as in the arbitration of agents contending 
for shared medium access.

\section{Background}
\label{sec:background}

\paragraph{Reinforcement Learning}
We consider a standard RL formulation based on Markov Decision
Process (MDP). An MDP is a tuple $<\cS, \cA, r, P, \gamma>$ where
$\cS$ and $\cA$ are the sets of states and actions, respectively, and
$\gamma \in [0, 1]$ is the discount factor. A transition probability
function $P : \cS \times \cA \rightarrow \cS$ maps states and actions
to a probability distribution over next states, and
$r : \cS\times\cA\rightarrow \real$ denotes the reward. The goal of RL
is to learn a policy $\pi: \cS \rightarrow \cA$ that solves the MDP by
maximizing the expected discounted return
$R_t =\expect{\sum_{k=0}^{\infty} \gamma^{k}r_{t+k}|\pi}$.  The policy
induces a value function $V^\pi(s)=\mathbb{E}_\pi[R_t|s_t=s]$, and an
action value function $Q^\pi(s,a)=\mathbb{E}_\pi[R_t|s_t=s,a_t=a]$.

\myparagraph{Actor-critic Method}
The main idea of the policy gradient method is to optimize the policy,
parametrized by $\theta^\pi$, in order to maximize the objective
$J(\theta) = \mathbb{E}_{s\sim p^\pi,a\sim \pi_\theta}[R]$ by directly
adjusting the parameters in the direction of the gradient. By the policy
gradient theorem \cite{sutton2000policy}, the gradient of the objective is:
\begin{align}
  \label{eq:pg}
\grad_\theta J(\pi_\theta) = \mathbb{E}_{s\sim \rho^\pi,a\sim \pi_\theta}[\grad_\theta \log \pi_\theta(a|s)Q^\pi(s, a)],
\end{align}
where $\rho^\pi$ is the state distribution.  Our baseline algorithmic framework is the {\em actor-critic} approach \cite{konda2003}. In this approach, an {\em actor} adjusts the parameters $\theta$ of the policy $\pi_\theta(s)$ by gradient ascent. Instead of the unknown true
action-value function $Q^\pi(s,a)$, its approximated version 
$Q^w(s,a)$ is used with parameter $w$. A {\em critic}
estimates the action-value function $Q^w(s,a)$ using an appropriate
policy evaluation algorithm such as temporal-difference learning \cite{tesauro1995temporal}. To reduce the variance of the gradient updates, some baseline function $b(s)$ is often subtracted from the action value, thereby resulting in $Q^\pi(s,a) - b(s)$ \cite{sutton1998}. A popular choice for this baseline function is the state value $V(s)$, which indicates the inherent ``goodness" of the state. This difference between the action value and the state value is often dubbed as the advantage $A(s,a)$ whose TD-error-based substitute $\delta_t=r_t+\gamma V(s_{t+1})-V(s_t)$ is an unbiased estimate of the advantage as in \cite{mnih2016a3c}. The actor-critic algorithm can also be applied to
a deterministic policy $\mu_\theta : \cS \rightarrow \cA$. By the deterministic policy gradient theorem \cite{dpg}, we update
the parameters as follows: 
\begin{align}
\label{eq:dpg}
  \grad_\theta J(\mu_\theta) = \mathbb{E}_{s\sim
  \rho^\mu}[\grad_\theta \mu_\theta(s)\grad_aQ^\mu(s,
  a)|_{a=\mu_\theta (s)}].
\end{align}

\myparagraph{MARL: Centralized Critic and Distributed Actor (CCDA)}

We formalize MARL using DEC-POMDP \citep{oliehoek2016concise},
which is a generalization of MDP to allow a distributed control by
multiple agents who may be incapable of observing the global state. A
DEC-POMDP is described by a tuple
$<\cS, \cA, r, P, \Omega, \cO, \gamma>$. We use bold-face fonts in some notations to highlight the context of multi-agents. Each agent
$i \in \set{N} $ chooses an action $a_i \in \cA,$ forming a joint action vector $\bm{a} = [a_i] \in \cA^n$  and has partial observations $o_i \in \Omega$ according to some
observation function
$\cO(\bm{s},i) : \cS \times \set{N} \mapsto \Omega.$ 
$P(\bm{s'}|\bm{s},\bm{a}): \cS \times \cA^n \mapsto [0,1]$ is the transition probability function. All agents share the same reward $r(\bm{s},\bm{u}): \cS \times \cA^n \mapsto \real.$
Each agent $i$
takes action $a_i$ based on its own policy $\pi^i(a_i|o_i)$.
As mentioned in Section~\ref{sec:intro}, our particular focus is on 
the centralized training and distributed
execution paradigm, where the actor-critic approach is a good fit to such a paradigm. Since the agents should execute in a distributed setting, 
each agent, say $i$, maintains its own actor that selects $i$'s action based only on what is partially observed by $i.$
The critic is naturally responsible for centralized training, and thus works in a centralized manner. Thus, the critic is allowed to have the global state
${\bm s}$ as its input, which includes all agents' observations and extra
information from the environment. The role of the critic is to ``criticize''
individual agent's actions. 
This centralized nature of the critic helps in 
providing more accurate feedback to the individual actors with
limited observation horizon.
In this case, each agent's policy, $\pi^i$, is updated by a variant of 
(\ref{eq:pg}) as:
\begin{align}
  \label{eq:mapg}
\grad_{\theta} J(\pi^i_{\theta}) = 
 \mathbb{E}_{\bm{s}\sim \rho^\pi,\bm{a}\sim
   \pi_{\theta}}[\grad_{\theta} \log \pi^i_\theta(a_i|o_i)(r+\gamma V(\bm{s}_{t+1})-V(\bm{s}_t))].
\end{align}

\section{Method}

\subsection{Communication Environment and Problem}
\label{sec:comm-env}

In practical scenarios where agents are typically separated but are
able to communicate over a shared medium, {\em e.g.}, a frequency channel in wireless communications, two important constraints are
imposed: bandwidth and contention for medium access \citep{wcp}. The
bandwidth constraint entails a limited amount of bits per unit time,
and the contention constraint involves having to avoid collision among
multiple transmissions due to the natural aspect of signal
broadcasting in wireless communication. Thus, only a restricted number
of agents are allowed to transmit their messages each time step for a reliable
message transfer. In this paper, we use a simple model to incorporate
that the aggregate information size per time step is limited by
$L_{\text{band}}$ bits and that only $K_{\text{sched}}$ out of $n$ agents
may broadcast their messages.

\myparagraph{Weight-based Scheduling} Noting that distributed
execution of agents is of significant importance, there may exist a
variety of scheduling mechanisms to schedule $K_{\text{sched}}$ agents
in a distributed manner. In this paper, we adopt a simple algorithm
that is weight-based, which we call WSA (Weight-based Scheduling
Algorithm). Once each agent decides its own weight, the
agents are scheduled based on their weights following a class of the
pre-defined rules.  We consider the following two specific ones among many
different proposals due to simplicity, but more importantly,
good approximation of wireless scheduling protocols in practice. 
\begin{compactitem}[$\circ$]
\item {\em Top($k$).} Selecting top $k$ agents in terms of their
  weight values.
\item {\em Softmax($k$).} Computing softmax values
  $\sigma_i({\bm w}) = \frac{e^{w_i}}{\sum_{j=1}^{n}e^{w_j}}$ for each
  agent $i,$ and then randomly selecting $k$ agents acoording to the probability distribution $[\sigma_i({\bm w})]_{i=1}^n.$
\end{compactitem}

Since distributed execution is one of our major operational
constraints in {\sched} or other CTDE-based MARL algorithms, Top($k$)
and Softmax($k$) should be realizable via a weight-based mechanism in
a distributed manner. In fact, this has been an active research topic
to date in wireless networking, where many algorithms exist
\citep{tassiulas1992stability, yy_scheduling,jiang2010distributed}.
Due to space limitation, we present how to obtain distributed versions
of those two rules based on weights in our supplementary material. To
summarize, using so-called CSMA (Carrier Sense Multiple Access) \citep{kurose2005computer},
which is a fully distributed MAC scheduler and forms a basis of Wi-Fi,
given agents' weight values, it is possible to implement Top($k$) and
Softmax($k$).

Our goal is to train agents so that every time each agent takes an action, only $K_{\text{sched}}$ agents can broadcast their messages with limited
size $L_{\text{band}}$ with the goal of receiving the highest cumulative reward via cooperation. Each agent should determine a policy described by its scheduling weights, encoded communication messages, and actions.  

\begin{wrapfigure}{R}{0.55\textwidth}
  \centering
  \includegraphics[width=0.53\columnwidth]{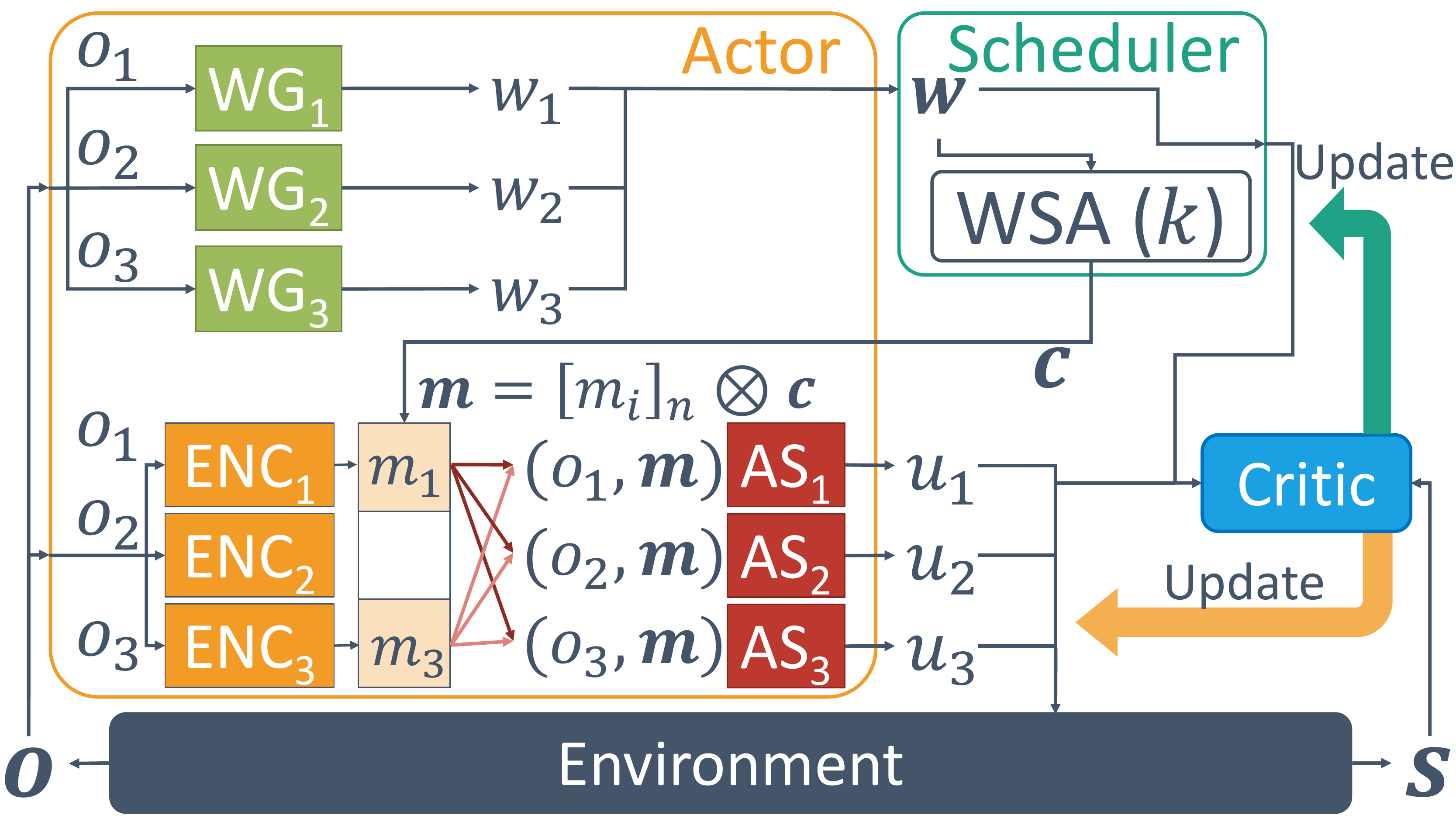}
  \vspace{-0.1cm}
\caption{\small Architecture of {\sched} with three agents. Agents 1 and 3 have been scheduled for this time step.}
  \label{fig:arch}
\vspace{-1.0cm}
\end{wrapfigure}

\subsection{Architecture}
\label{sec:architecture}

To this end, we propose a new deep MARL framework with scheduled communications, 
called {\sched}, whose overall architecture is depicted in 
Figure~\ref{fig:arch}. {\sched} consists of the following three components: 
{\em (i)} actor network, {\em (ii)} scheduler, and
{\em (iii)} critic network.
This section is devoted to presenting the architecture only, whose details are presented in the subsequent sections.

\myparagraph{Neural networks} 
The actor network is the collection of $n$ per-agent individual actor networks, where each agent $i$'s individual actor network consists of a triple of the following networks: a \linebreak

message encoder, an action selector, and a weight generator, as specified by: 
\begin{align*}
&\text{\bf message encoder} \  f^i_{\text{enc}}: o_i \mapsto m_i, \cr \quad
&\text{\bf action selector} \  f_{\text{as}}^i: (o_i, \vec{m} \otimes \vec{c}) \mapsto u_i, \cr
&\text{\hspace{0cm}} \text{\bf weight generator} \  f_{\text{wg}}^i: o_i \mapsto w_i. 
\end{align*}
Here, $\vec{m} = [m_i]_n$\footnote{We use $[\cdot]_n$ to mean the $n$-dimensional vector, where $n$ is the number of agents.} is the vector of each $i$'s encoded message $m_i.$ An agent schedule vector $\vec{c} = [c_i]_n,$ $c_i \in \{0,1\}$ represents whether each agent is scheduled.
Note that agent $i$'s encoded message
$m_i$ is generated by a neural network $f^i_{\text{enc}}: o_i \mapsto m_i.$ The operator ``$\otimes$'' concatenates all the scheduled agents' messages. For example, 
for $\vec{m} = [010,111,101]$ and $\vec{c} = [110],$ $\vec{m} \otimes \vec{c} = 010111.$ This concatenation with the schedule profile $\vec{c}$ means that 
only those agents scheduled in $\vec{c}$ may broadcast their messages to all other agents. We denote by $\theta^i_{\text{as}}$, $\theta^i_{\text{wg}}$, and $\theta^i_{\text{enc}}$ the parameters of the action selector, the weight generator, and the encoder of agent $i$, respectively, where we let $\bm{\theta}_{\text{as}} = [\theta^i_{\text{as}}]_n$, and similarly define $\bm{\theta}_{\text{wg}}$ and $\bm{\theta}_{\text{enc}}$.

\myparagraph{Coupling: Actor and Scheduler} 
Encoder, weight generator and the scheduler are the modules for handling the constraints of
limited bandwidth and shared medium access. Their common goal is to
learn the state-dependent ``importance'' of individual agent's
observation, encoders for generating compressed messages and the scheduler for
being used as a basis of an external scheduling mechanism based on the weights generated by per-agent weight generators. These three
modules work together to smartly respond to time-varying states. The
action selector is trained to decode the incoming message, and
consequently, to take a good action for maximizing the reward. At
every time step, the schedule profile ${\bm c}$ varies depending on
the observation of each agent, so the incoming message $\vec{m}$ comes
from a different combination of agents. Since the agents can be
heterogeneous and they have their own encoder, the action selector
must be able to make sense of incoming messages from different
senders. However, the weight generator's policy changes, 
the distribution of incoming messages also changes, which is in turn affected by the pre-defined WSA. Thus, the action
selector should adjust to this changed scheduling. This also affects
the encoder in turn. The updates of the encoder and the action
selector trigger the update of the scheduler again. Hence, 
weight generators, message encoders, and action selectors  
are strongly coupled with dependence on a specific WSA, and we train those three networks at the
same time with a common critic.

\myparagraph{Scheduling logic} The schedule profile $\vec{c}$ is
determined by the WSA module, which is mathematically a mapping from all
agents' weights $\vec{w}$ (generated by $f^i_{\text{wg}}$) to
$\vec{c}.$ Typical examples of these mappings are {\em Top($k$)} and
{\em Softmax($k$)}, as mentioned above. The scheduler of each agent
is trained appropriately depending on the employed WSA algorithm.

\subsection{Training and Execution}
\label{sec:act}

\begin{wrapfigure}{R}{0.3\textwidth}
\vspace{-0.4cm}
  \centering
  \includegraphics[width=0.3\columnwidth]{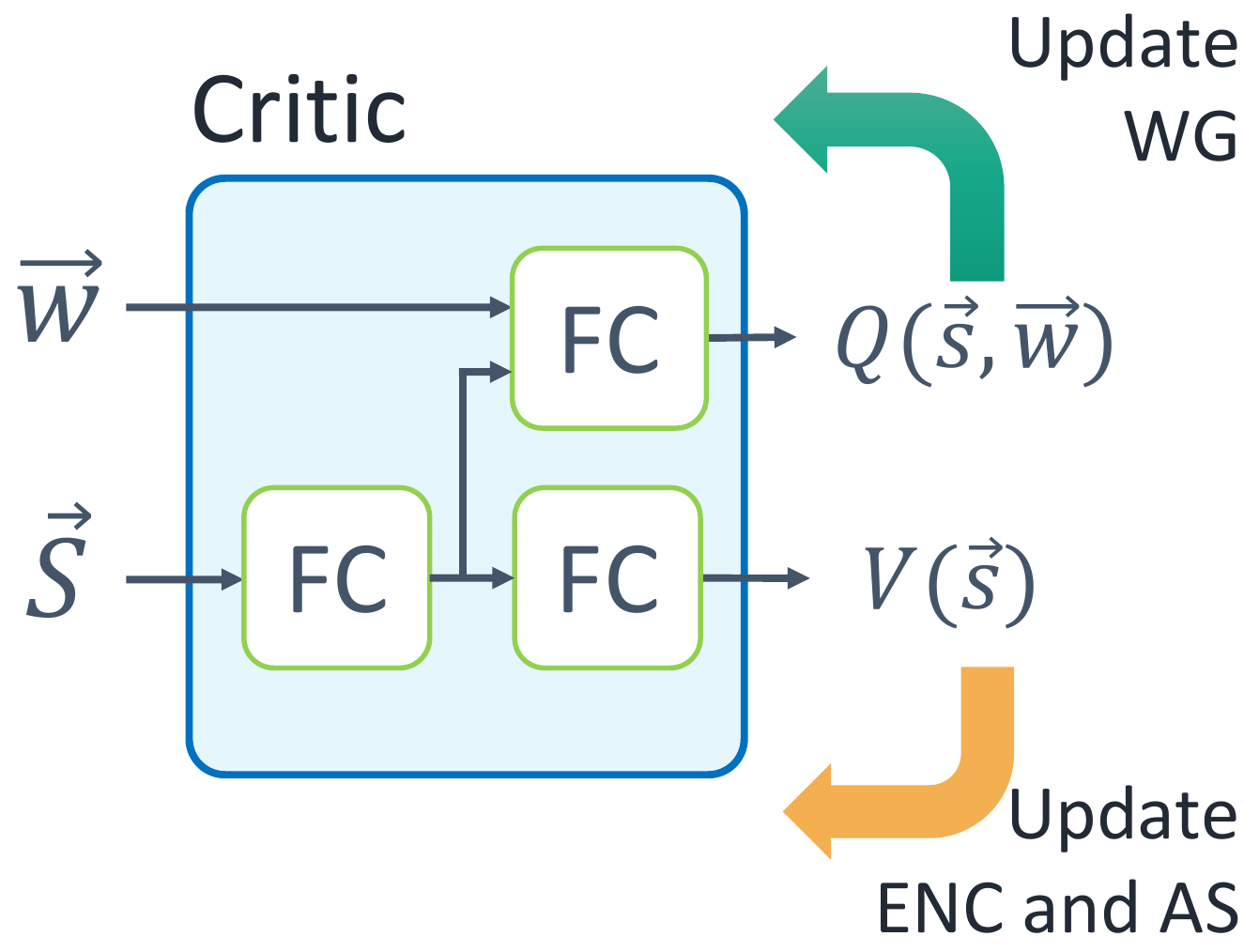}
\caption{\small Architecture of the critic. FC stands for fully connected
  neural network.}
  \label{fig:critic}
\vspace{-0.6cm}
\end{wrapfigure}

In the centralized training with distributed execution, for a given
WSA, we include all components and modules in Figure~\ref{fig:arch} to
search for
$\bm{\theta}_{\text{as}}$, $\bm{\theta}_{\text{wg}}$, and
$\bm{\theta}_{\text{enc}}$, whereas in execution, each agent $i$ runs 
a certain shared medium access mechanism, well-modeled by a weight-based scheduler, and just needs three agent-specific
parameters $\theta^i_{\text{as}}$, $\theta^i_{\text{wg}}$, and
$\theta^i_{\text{enc}}$.

\subsubsection{Centralized Training}
\label{sec:training}

\paragraph{Centralized critic}
The actor is trained by dividing it into two parts: {\em (i)} message encoders and action selectors, and {\em (ii)} weight generators. This partitioning is motivated by the fact that it is hard to update both parts with one backpropagation since WSA is not differentiable. 
To update the actor, we use a centralized critic parametrized by
$\theta_{\text{c}}$ to
estimate the state value function $V_{{\theta}_{\text{c}}}(\bm{s})$ for
the action selectors and message encoders, and the action-value function
$Q_{{\theta}_{\text{c}}}^\pi(\bm{s},\bm{w})$ for the weight generators. 
The critic is used only when training, and it can
use the global state $\bm{s}$, which includes the observation of all agents.
All networks in the actor are trained with gradient-based on temporal difference
backups. To share common features between
$V_{{\theta}_{\text{c}}}(\bm{s})$ and
$Q_{{\theta}_{\text{c}}}^\pi(\bm{s},\bm{w})$ and perform
efficient training, we use shared parameters in the lower layers of the
neural network between the two functions, as shown in Figure \ref{fig:critic}.

\myparagraph{Weight generators}
We consider the collection of all agents' WGs as a single neural network $\mu_{\bm{\theta}_\text{wg}}(\bm{o})$ mapping from $\bm{o}$ to $\bm{w}$,
parametrized by $\bm{\theta}_\text{wg}$. 
Noting that $w_i$ is a continuous value, we apply 
the DDPG algorithm \citep{ddpg}, where the entire policy gradient of the collection of WGs is given by:
\begin{align*}
\grad_{\bm{\theta}_{\text{wg}}} J( \bm{\theta}_{\text{wg}} , \cdot ) = \mathbb{E}_{\bm{w} \sim \mu_{\bm{\theta}_\text{wg}}} [
\grad_{\bm{\theta}_{\text{wg}}} \mu_{\bm{\theta}_\text{wg}} ({\bm o}) \grad_{{\bm w}} Q_{{\theta}_{\text{c}}} ({\bm s}, {\bm w})
|_{{\bm w} = \mu_{\bm{\theta}_\text{wg}} ({\bm o})}]. \nonumber
\end{align*}
We sample the policy gradient for sufficient amount of experience in
the set of all scheduling profiles, {\em i.e.}, $\set{C} = \{\vec{c} \mid \sum_{c_i} \leq k \}$.
The values of $Q_{{\theta}_{\text{c}}} ({\bm s}, {\bm w})$ are estimated by the
centralized critic, where ${\bm s}$ is the global state corresponding to
${\bm o}$ in a sample.

\myparagraph{Message encoders and action selectors}
The observation of each agent travels through the encoder and the
action selector. We thus serialize $f^i_{\text{enc}}$ and $f^i_{\text{as}}$ together and
merge the encoders and actions selectors of all agents into one aggregate network
$\pi_{\bm{\theta}_{\text{u}}}({\bm u}|{\bm o},{\bm c})$, which is parametrized by $\bm{\theta}_{\text{u}}=\{\bm{\theta}_{\text{enc}},\bm{\theta}_{\text{as}}\}$. This
aggregate network $\pi_{\bm{\theta}_{\text{u}}}$ learns via backpropagation of actor-critic
policy gradients, described below. The gradient of this objective
function, which is a variant of (\ref{eq:mapg}), is given by
\begin{align}
  \label{eq:a_gradient}
  \grad_{\bm{\theta}_{\text{u}}} J(\cdot,\bm{\theta}_{\text{u}}) 
  &= \mathbb{E}_{\bm{s}\sim \rho^\pi,\bm{u}\sim \pi_{\bm{\theta}_{\text{u}}}}[\grad_{\bm{\theta}_{\text{u}}} \log
                                       \pi_{\bm{\theta}_{\text{u}}}({\bm u}|{\bm o}, {\bm c})[r + \gamma V_{{\theta}_{\text{c}}}({\bm s'}) - V_{{\theta}_{\text{c}}}({\bm s})]],
\end{align}
where ${\bm s}$ and ${\bm s'}$ are the global states corresponding to
the observations at current and next time step. We can get the value
of state $V_{{\theta}_{\text{c}}}({\bm s})$ from the centralized critic and then adjust the
parameters $\bm{\theta}_{\text{u}}$ via gradient ascent accordingly.

\subsubsection{Distributed Execution}
\label{sec:execution}

 In execution, each agent $i$ should be able to determine the
  scheduling weight $w_i$, encoded message $m_i$, and action selection
  $u_i$ in a distributed manner. This process must be based on its own observation, and the weights generated by 
  its own action selector, message encoder, and weight
  generator with the parameters $\theta^i_{\text{as}}$, $\theta^i_{\text{enc}}$,     and $\theta^i_{\text{wg}}$, respectively.
  After each agent determines its scheduling
  weight, $K_{\text{sched}}$ agents are scheduled by WSA, which leads the encoded
  messages of scheduled agents to be broadcast to all agents. Finally, each agent
  finally selects an action by using received messages. This process is
  sequentially repeated under different observations over time.

\section{Experiment}
\label{sec:exp}

\paragraph{Environments} To evaluate \sched\footnote{The code is available on \texttt{https://github.com/rhoowd/sched\_net}}, we consider two different
environments for demonstrative purposes: Predator and Prey (PP) which
is used in \citet{stone2000multiagent}, and Cooperative Communication
and Navigation (CCN) which is the simplified version of the one in
\citet{lowe2017multi}. The detailed experimental environments are
elaborated in the following subsections as well as in supplementary material.
We take the communication environment into our consideration as
follows. $k$ out of all agents can have the chance to broadcast
the message whose bandwidth\footnote{The unit of bandwidth is 2 bytes
  which can express one real value (float16 type)} is limited by $l$.

\myparagraph{Tested algorithms and setup}
We perform experiments in aforementioned environments. We compare \sched~with a variant of
DIAL,\footnote{We train and execute DIAL without discretize/regularize
  unit (DRU), because in our setting, agents can exchange messages that
  can express real values.}~\citep{dial} which allows communication
with limited bandwidth. During the execution of DIAL, the limited
number ($k$) of agents are scheduled following a simple
round robin scheduling algorithm, and the agent reuses the
outdated messages of non-scheduled agents to make a decision on the action to take,
which is called DIAL($k$). The other baselines are independent DQN
(IDQN)~\citep{tampuu2017multiagent} and COMA~\citep{coma} in which no
agent is allowed to communicate. To see the impact of scheduling in
\sched, we compare \sched~with {\em (i)} RR (round robin), which is a
canonical scheduling method in communication systems where all agents
are sequentially scheduled, and {\em (ii)} FC (full communication), which is
the ideal configuration, wherein all the agents can send their
messages without any scheduling or bandwidth constraints. We also
diversify the WSA in \sched~into: {\em (i)} Sched-Softmax(1) and {\em (ii)}
Sched-Top(1) whose details are in Section~\ref{sec:comm-env}.
We train our models until convergence, and then evaluate them by
averaging metrics for 1,000 iterations. The shaded area in each plot
denotes 95\% confidence intervals based on 6-10 runs with 
different seeds.

\begin{figure}[t!]
\captionsetup[subfigure]{justification=centering}
  \begin{center}
\hspace{-0.2cm}
   \subfloat[\scriptsize PP: Comparison with other baselines. $k=1, l=2$] {\label{fig:p_pp}
      \includegraphics[width=0.32\columnwidth]{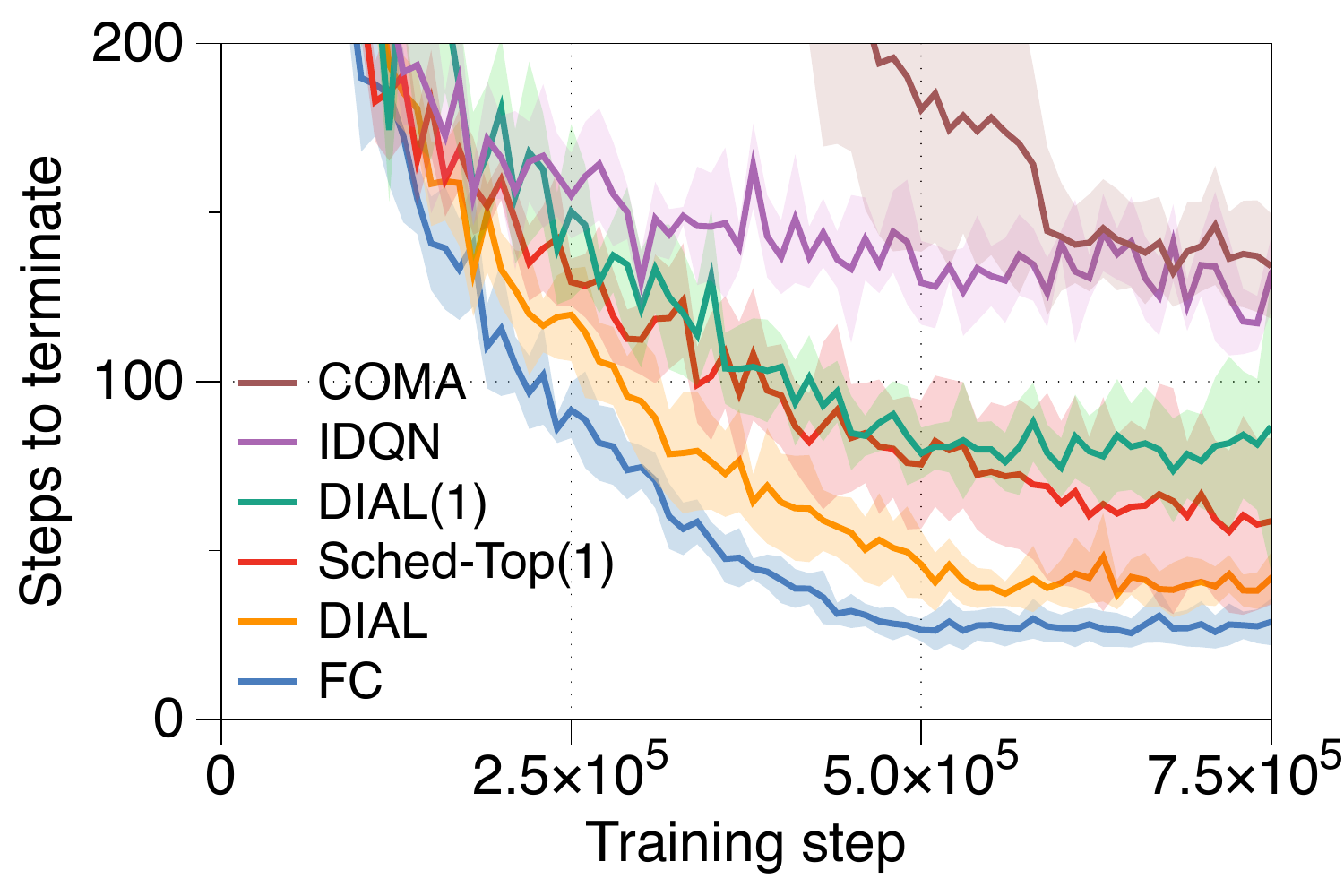} }
      \hspace{-0.2cm}    
    \subfloat[\scriptsize PP: Comparison with scheduling schemes. $l=2$]{\label{fig:pp_schedule}
      \includegraphics[width=0.32\columnwidth]{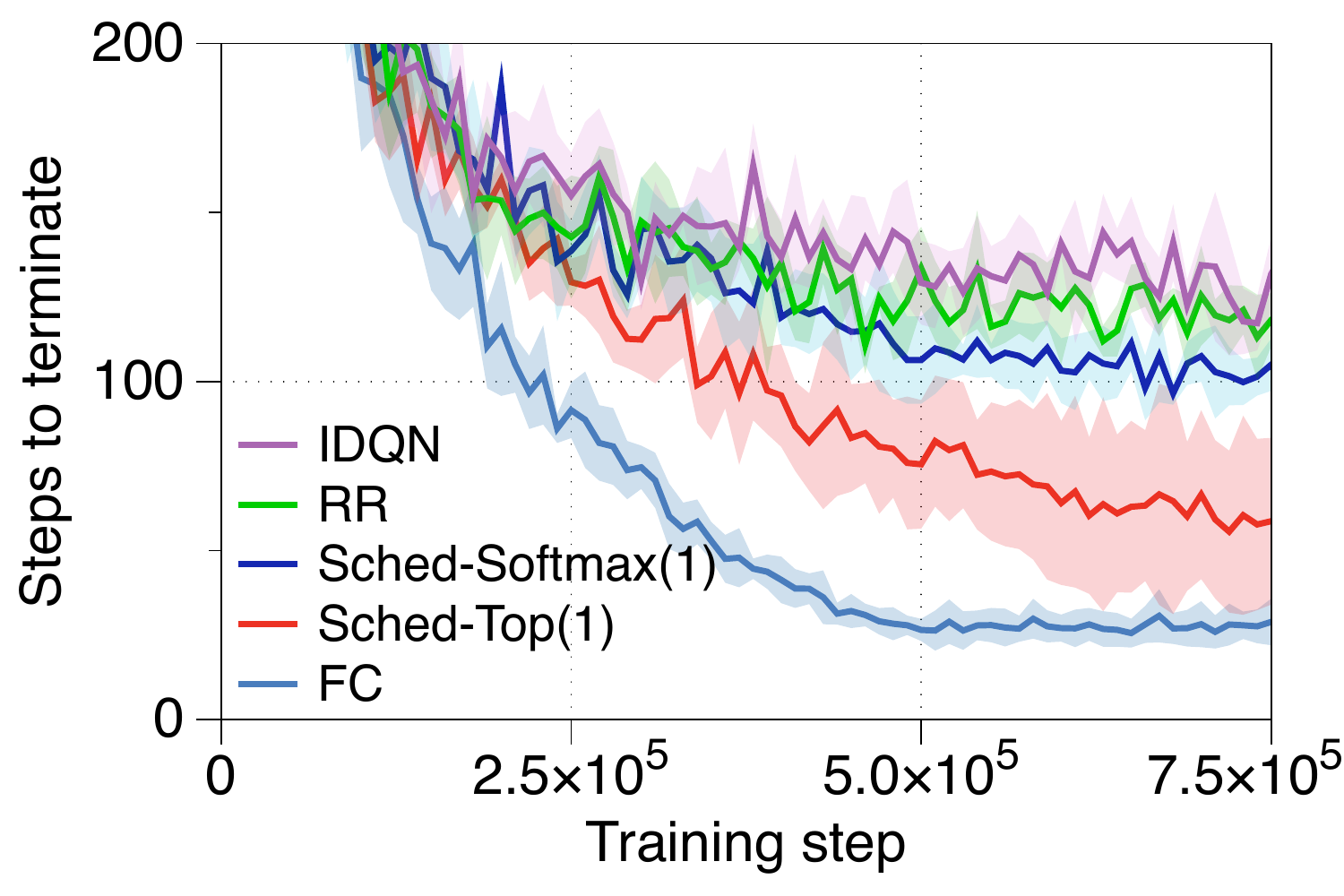} }
    \hspace{-0.2cm} 
    \subfloat[\scriptsize CCN:  Comparison with scheduling schemes. $k=1, l=1$]
    {\label{fig:ccn_p}
     \includegraphics[width=0.32\columnwidth]{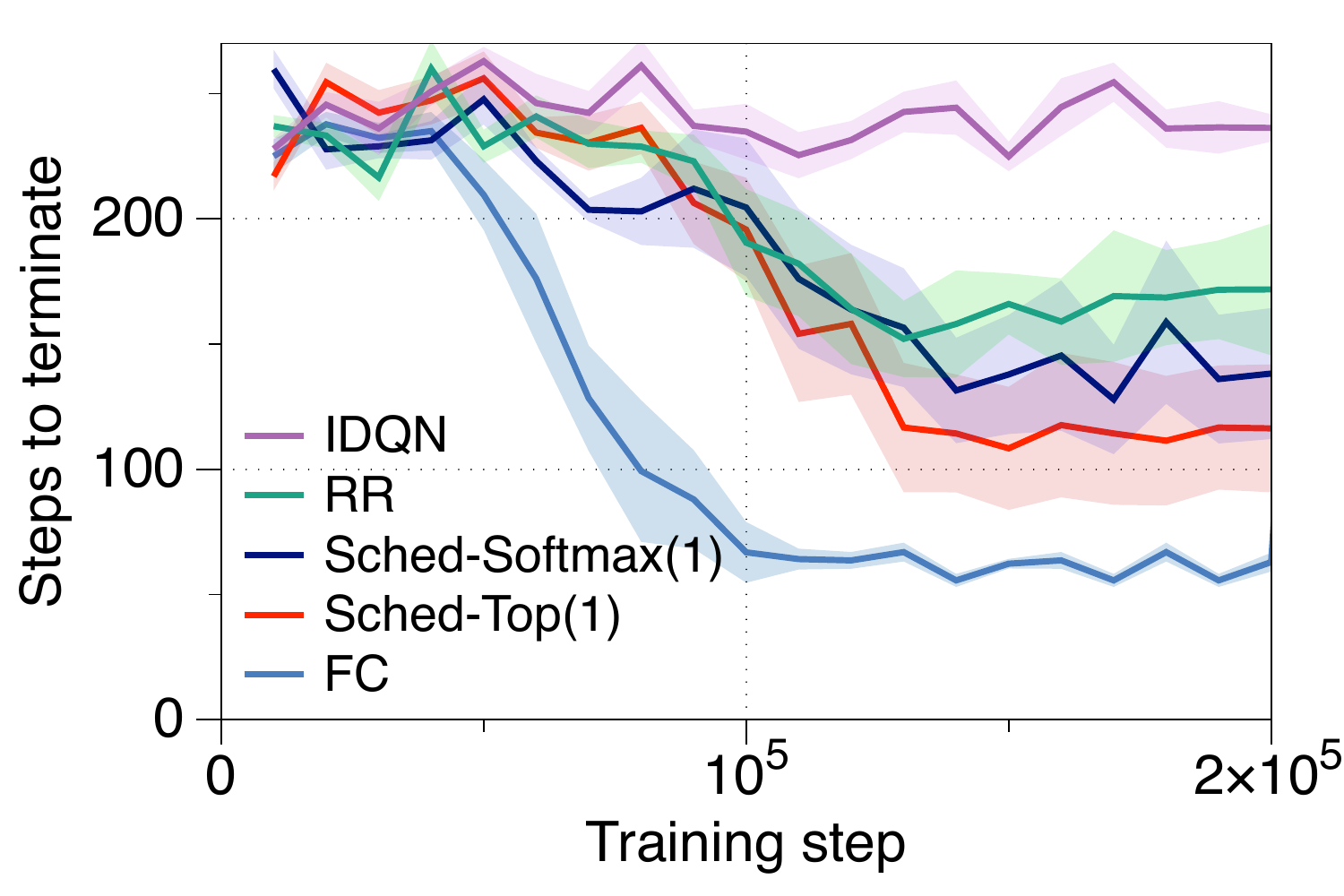} }
\end{center}
\vspace{-0.4cm}
\caption{\small Learning curves during the learning of the PP and CCN tasks. The plots
  show the average time taken to complete the task, where shorter time
  is better for the agents.}
\label{fig:result}
\vspace{-0.4cm}
\end{figure}

\subsection{Predator and Prey}
\label{sec:pp}

In this task, there are multiple agents who must capture a randomly
moving prey. Agents' observations include position of themselves and
the relative positions of prey, if observed. We employ four
agents, and they have different observation horizons, where only agent 1
has a $5 \times 5$ view while agents 2, 3, and 4 have a smaller,
$3 \times 3$ view. The predators are rewarded when they capture the
prey, and thus the performance metric is the number of time steps
taken to capture the prey.

\myparagraph{Result in PP} Figure \ref{fig:p_pp} illustrates the
learning curve of 750,000 steps in PP. In FC, since the agents can use
full state information even during execution, they achieve the best
performance.  \sched~outperforms IDQN and COMA in which
communication is not allowed. It is observed that agents first find the prey, and then follow it until all other agents also eventually observe the prey. An
agent successfully learns to follow the prey after it observes the prey
but that it takes a long time to meet the prey for the first time. If the
agent broadcasts a message that includes the location information of the
prey, then other agents can find the prey more quickly. Thus, it is
natural that \sched~and DIAL perform better than IDQN or COMA, because
they are trained to work with communication. However, DIAL is not
trained for working under medium contention constraints. Although DIAL works
well when there is no contention constraints, under the condition
where only one agent is scheduled to broadcast the message by a simple
scheduling algorithm ({\em i.e.}, RR),
the average number of steps to capture the prey in DIAL(1) is larger
than that of \sched -Top(1), because the outdated messages of
non-scheduled agents is noisy for the agents to decide on
actions. Thus, we should consider the scheduling from when we train
the agents to make them work in a demanding environment.

\begin{wrapfigure}{R}{0.36\textwidth}
  \vspace{-0.5cm}
  \includegraphics[width=0.35\columnwidth]{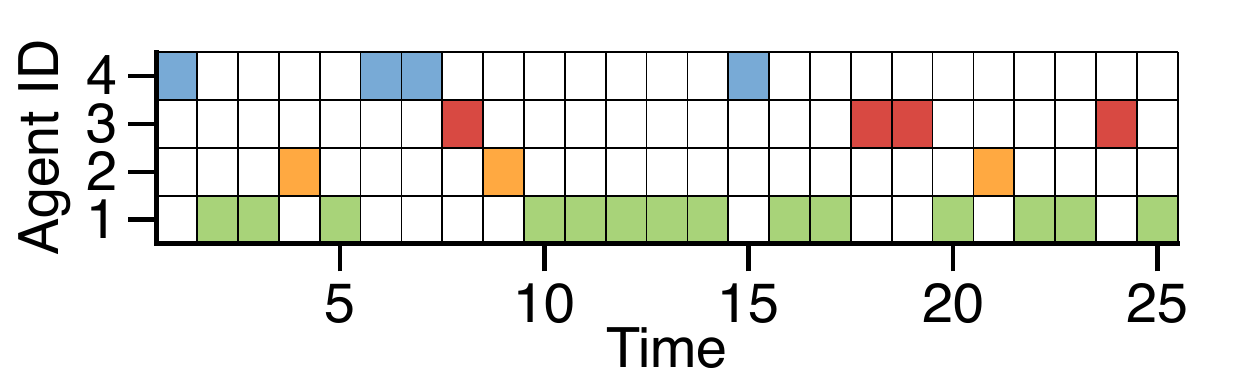}
  \vspace{-0.3cm} \captionof{figure}{\small Instances of scheduling results over 25 time steps in  PP}
  \label{fig:sched}
  \vspace{0.3cm}                
  \includegraphics[width=0.33\columnwidth]{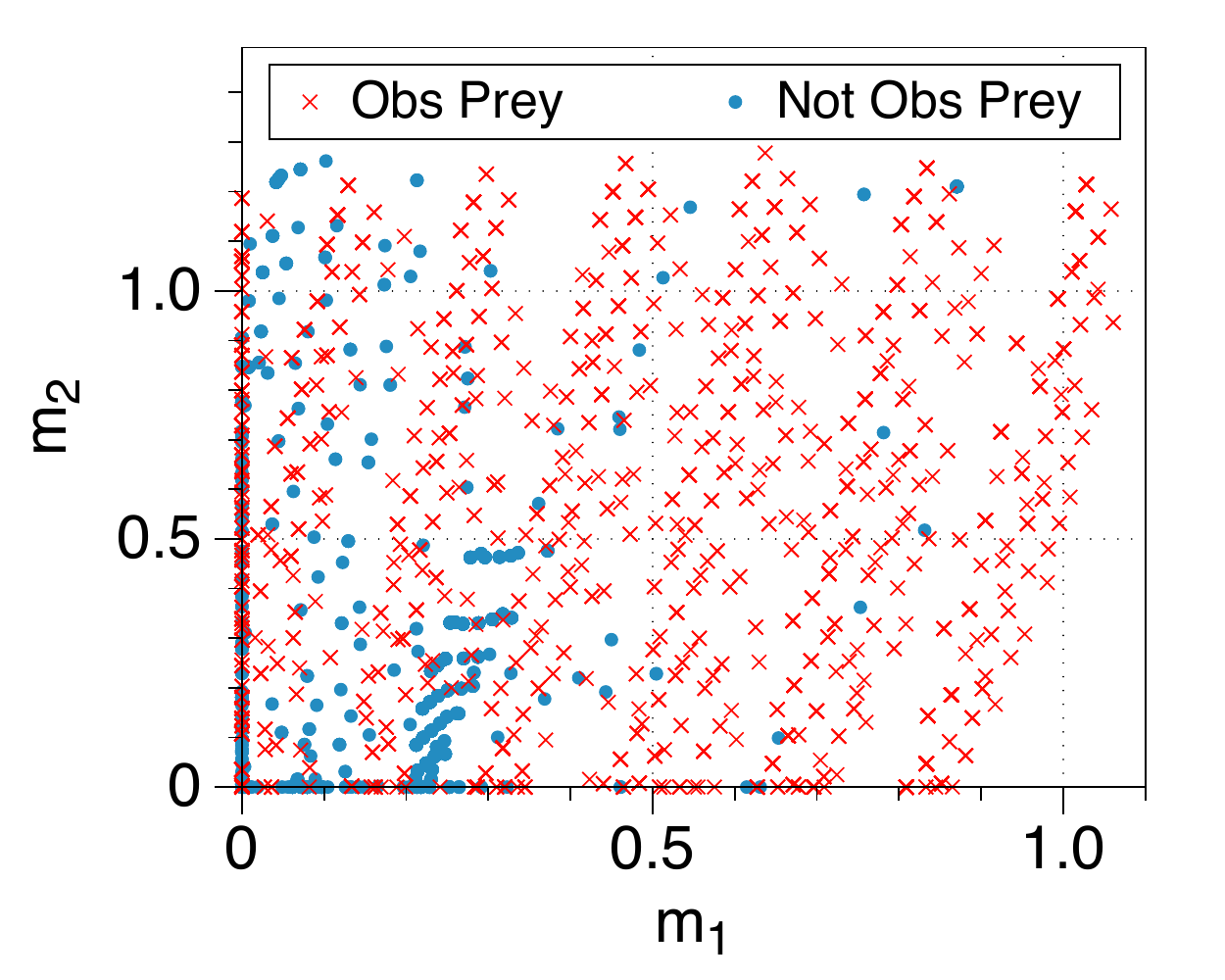}
  \vspace{-0.4cm} \captionof{figure}{\small Encoded messages projected onto 2D plane in PP task}
  \label{fig:msg}
  \vspace{-0.9cm}
\end{wrapfigure}

\myparagraph{Impact of intelligent scheduling} In Figure
\ref{fig:pp_schedule}, we observe that IDQN, RR, and \sched
-Softmax(1) lie more or less on a comparable performance tier, with
\sched -Softmax(1) as the best in the tier.  \sched -Top(1)
demonstrates a non-negligible gap better than the said tier, implying
that a deterministic selection improves the agents' collective rewards
the best. In particular, \sched -Top(1) improves the performance by
43\% compared to RR. Figure \ref{fig:pp_schedule} lets us infer that,
while all the agents are trained under the same conditions except for
the scheduler, the difference in the scheduler is the sole determining
factor for the variation in the performance levels. Thus, ablating away
the benefit from smart encoding, the intelligent scheduling element in
\sched~can be accredited with the better performance.

\myparagraph{Weight-based Scheduling}
We attempt to explain the internal behavior of \sched~by investigating
instances of temporal scheduling profiles obtained during the
execution.  We observe that \sched~has learned to schedule those agents
with a farther observation horizon, realizing the rationale of
importance-based assignment of scheduling priority also for the PP
scenario. Recall that Agent 1 has a wider view and thus tends to
obtain valuable observation more frequently. In
Figure~\ref{fig:sched}, we see that scheduling chances are distributed
over (14, 3, 4, 4) where corresponding average 
weights are (0.74,
0.27, 0.26, 0.26), implying that those with greater observation power
tend to be scheduled more often.

\myparagraph{Message encoding}

We now attempt to understand what the predator agents communicate when
performing the task. Figure~\ref{fig:msg} shows the projections of the
messages onto a 2D plane, which is generated by the scheduled agent
under \sched -Top(1) with $l=2$. When the agent does not observe the
prey (blue circle in Figure), most of the messages reside in the
bottom or the left partition of the plot. On the other hand, the
messages have large variance when it observes the prey (red `x'). This
is because the agent should transfer more informative messages that
implicitly include the location of the prey, when it observes the
prey. Further analysis of the messages is presented in our
supplementary material.

\subsection{Cooperative communication and navigation}
In this task, each agent's goal is to arrive at a pre-specified
destination on its one-dimensional world, and they collect a joint
reward when both agents reach their respective destination. Each agent has a zero
observation horizon around itself, but it can observe the situation of
the other agent. We introduce heterogeneity into the scenario, where
the agent-destination distance at the beginning of the task differs
across agents. The metric used to gauge the performance is the number
of time steps taken to complete the CCN task.

\myparagraph{Result in CCN} We examine the CCN environment whose results
are shown in Figure \ref{fig:ccn_p}.  \sched~and other baselines were
trained for 200,000 steps. As expected, IDQN takes the longest time,
and FC takes the shortest time. RR exhibits mediocre performance,
better than IDQN, because agents at least take turns in obtaining the
communication opportunity. Of particular interest is \sched,
outperforming both IDQN and RR with a non-negligible gap. We remark
that the deterministic selection with \sched -Top(1) slightly beats
the probabilistic counterpart, \sched -Softmax(1). The 32\% improved
gap between RR and \sched~clearly portrays the effects of intelligent
scheduling, as the carefully learned scheduling method of \sched~was
shown to complete the CCN task faster than the simplistic RR.

\begin{wrapfigure}{R}{0.35\textwidth}
  \includegraphics[width=0.35\columnwidth]{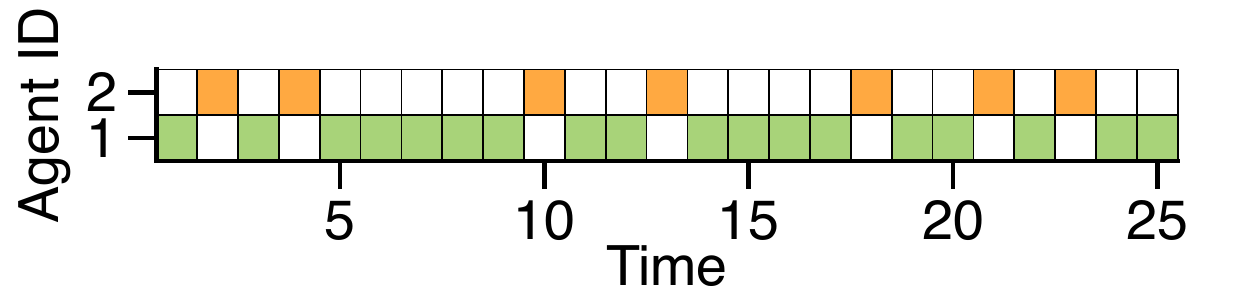}
  \vspace{-0.1cm} \captionof{figure}{\small Instances of scheduling results
    over 25 time steps in  CCN}
  \label{fig:sched_ccn}
\end{wrapfigure}

\myparagraph{Scheduling in CCN}
As Agent 2 is farther from its destination than Agent 1, we observe
that Agent 1 is scheduled more frequently to drive Agent 2 to its
destination (7 vs. 18), as shown in Figure~\ref{fig:sched_ccn}. This
evidences that \sched~flexibly adapts to heterogeneity of agents via
scheduling. Towards more efficient completion of the task, a rationale
of {\em more scheduling for more important agents} should be
implemented.
This is in accordance with the results obtained from PP environments:
more important agents are scheduled more.

\section{Conclusion}
\label{sec:discussion}

We have proposed \sched~for learning to schedule inter-agent communications
in fully-cooperative multi-agent tasks. In \sched, we have the centralized critic
giving feedback to the actor, which consists of message encoders, action selectors, and weight generators of each individual agent. The message encoders and action selectors are criticized
towards compressing observations more efficiently and selecting actions
that are more rewarding in view of the cooperative task at hand. Meanwhile,
the weight generators are criticized such that $k$ agents with apparently more valuable observation are allowed to access the shared medium and broadcast their messages to all other agents. Empirical results and an accompanying ablation study indicate that the learnt encoding and scheduling behavior each significantly improve the agents' performance. We have observed that an
intelligent, distributed  communication scheduling can aid in a more efficient, coordinated,
and rewarding behavior of learning agents in the MARL setting.


\newpage
\appendix
\addcontentsline{toc}{section}{Appendices}
\section*{Supplementary Material}

\section{\sched~Training Algorithm}

The training algorithm for {\sched} is provided in Algorithm~\ref{algo:schednet}. The parameters of the message encoder are assumed to be included in the actor network. Thus, we use the notation $f^i_\text{as}(o^i, \vec{c}) = f^i_\text{as}(o^i, f^i_\text{enc}(o^i) \otimes \vec{c})$ to simplify the presentation.




\begin{algorithm}[h!]
\caption{\sched}
\label{algo:schednet}
\begin{algorithmic}[1]
\State Initialize actor parameters $\theta_u$, scheduler parameters $\theta_\text{wg}$, and critic parameters $\theta_c$
\State Initialize target scheduler parameters $\theta'_\text{wg}$, and target critic parameters $\theta'_{c}$
\For{episode = 1 to $M$}
	\State Observe initial state $\bm{s}$
    \For{$t = 1$ to $T$}
        \State $\bm{w}_t \gets$ the priority $w^i = f^i_\text{wg}(o^i)$ of each agent $i$
        \State Get schedule $\vec{c_t}$ from $\bm{w}_t$ 
        \State $\bm{u}_t \gets$ the action $u^i = f^i_\text{as}(o^i,   \vec{c_t})$ of each agent $i$  
        \State Execute the actions $\bm{u}_t$ and observe the reward $r_t$ and next state $\bm{s}_{t+1}$
        \State Store ($\bm{s}_t$, $\bm{u}_t$, $r_t$, $\bm{s}_{t+1}$, $\vec{c_t}$, $\bm{w}_t$) in the replay buffer $B$
        \State Sample a minibatch of $S$ samples ($\bm{s}_k$, $\bm{u}_k$, $r_k$, $\bm{s}_{k+1}$, $\vec{c_k}$, $\bm{w}_k$) from $B$
        \State Set $y_k = r_k + \gamma \bar V(\bm{s}_{k+1})$
        \State Set $\hat{y}_k = r_k + \gamma \bar Q(\bm{s}_{k+1}, \bar{f}_\text{wg}^i(\bm{o}_{k+1}, \vec{c_{k+1}}))$
        \State Update the critic by minimizing the loss: 
        \begin{equation*}
        	L  = \frac{1}{S}\sum_k ((y_k - V(\bm{s}_k))^2 + (\hat{y}_k - Q(\bm{s},\bm{w}_k))^2)
        \end{equation*}
        \State Update the actor along with the encoder using sampled policy gradient:
        \begin{equation*}
        \grad_{\bm{\theta}_{\text{u}}} J(\cdot,\bm{\theta}_{\text{u}}) = \mathbb{E}_{\bm{s}\sim \rho^\pi,\bm{u}\sim \pi}[\grad_{\bm{\theta}_{\text{u}}} \log \pi({\bm u}|{\bm o}, {\bm c})[r + \gamma V_{{\theta}_{\text{c}}}({\bm s'}) - V_{{\theta}_{\text{c}}}({\bm s})]]
        \end{equation*}
        \State Update scheduler using sampled policy gradient:
        \begin{equation*}
       	\grad_{\bm{\theta}_{\text{wg}}} J( \bm{\theta}_{\text{wg}} , \cdot ) = \mathbb{E}_{\bm{w} \sim \mu} [
\grad_{\bm{\theta}_{\text{wg}}} \mu ({\bm o}) \grad_{{\bm w}} Q_{{\theta}_{\text{c}}} ({\bm s}, {\bm w})
|_{{\bm w} = \mu ({\bm o})} ] 
        \end{equation*}
        \State Update target network parameters:
        \begin{equation*}
        	\theta'_\text{wg} \gets \tau \theta_\text{wg} + (1-\tau)\theta'_\text{wg}
        \end{equation*}
        \begin{equation*}
        	\theta'_{c} \gets \tau \theta_{c} + (1-\tau)\theta'_{c}
        \end{equation*}
    \EndFor
\EndFor
\end{algorithmic}
\end{algorithm}

\section{Details of Environments and Implementation}

\begin{figure}[h!]
  \begin{center}
    \subfloat[Predator and prey (PP)]
    {\label{fig:pp}
      \includegraphics[width=0.40\columnwidth]{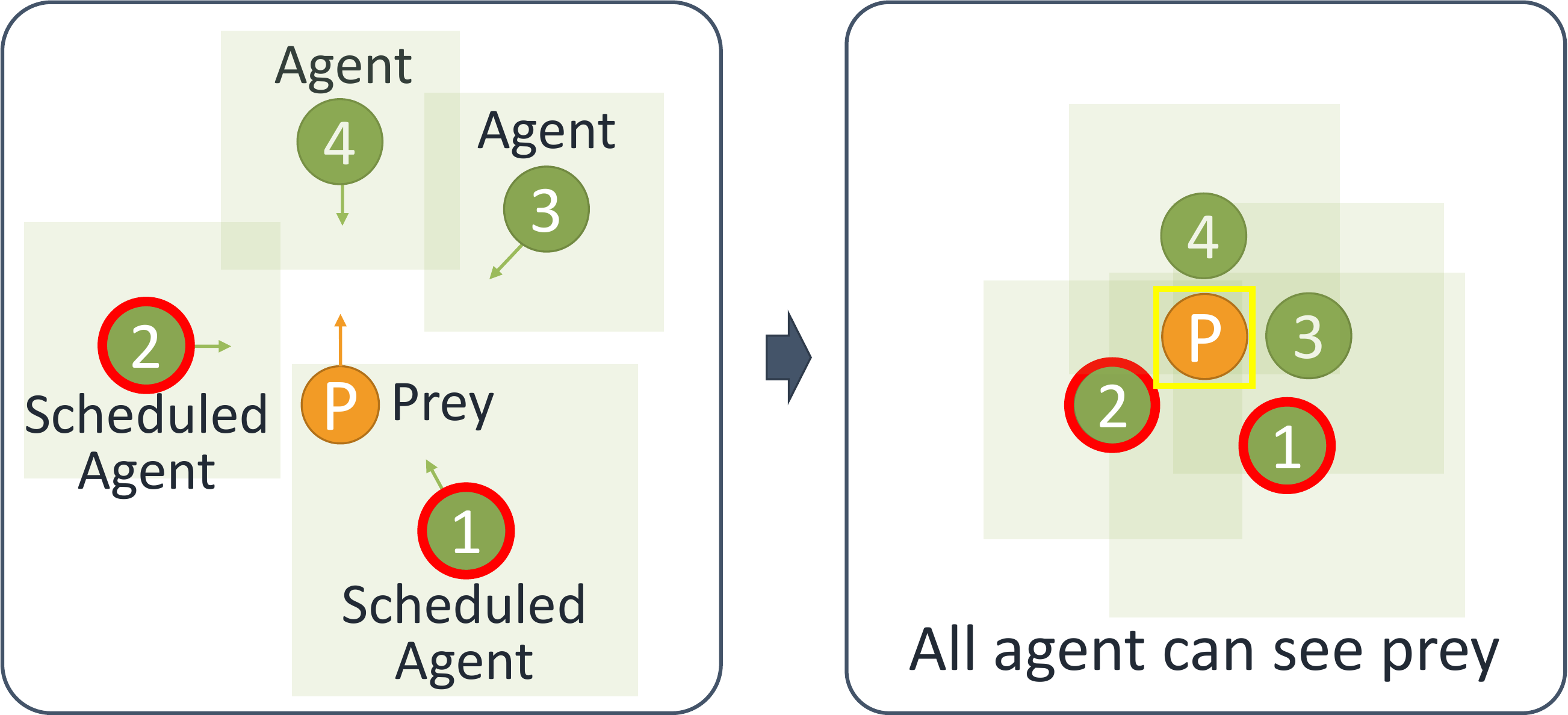} }
    \hspace{0.7cm}
    \subfloat[Cooperative communication and navigation (CCN)] {\label{fig:ccn}
      \includegraphics[width=0.44\columnwidth]{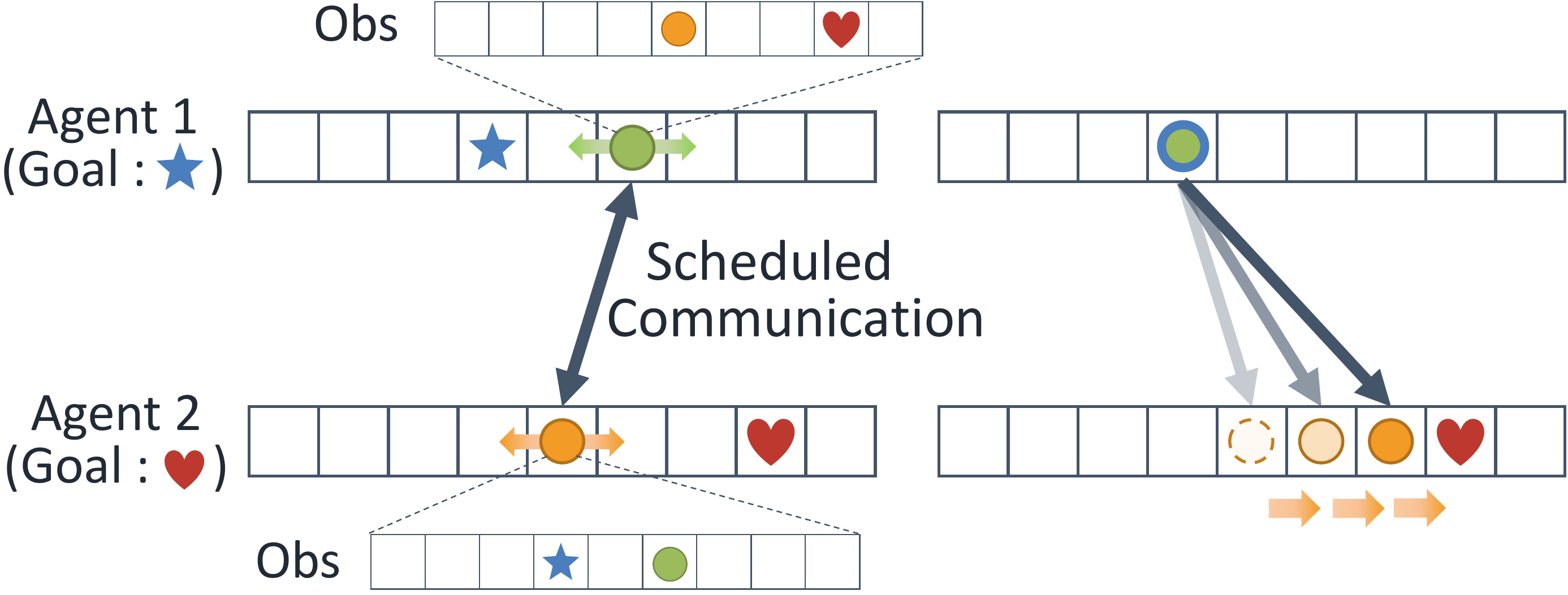} }
  \end{center}
  \vspace{-0.2cm}
  \caption{Illustrations of the experimental environment}
  \label{fig:scenario}
\end{figure}

\subsection{Environments: PP and CCN}

\myparagraph{Predator and prey}
We assess \sched~in this predator-prey setting as in
\cite{stone2000multiagent}, illustrated in Figure \ref{fig:pp}.  This
setting involves a discretized grid world and multiple cooperating
predators who must capture a randomly moving prey. Agents'
observations include position of themselves and the relative positions
of the prey, if observed. The observation horizon of each predator is
limited, thereby emphasizing the need for communication. The
termination criterion for the task is that all agents observe the
prey, as in the right of Figure~\ref{fig:pp}. The predators are
rewarded when the task is terminated.
We note that agents may be endowed with
different observation horizons, making them heterogeneous. 
We employ four agents in our experiment, where only agent 1 has a $5 \times 5$ view
while agents 2, 3, and 4 have a smaller, $3 \times 3$ view.
The performance metric is the number of time steps taken to capture the prey.

\paragraph{Cooperative communication and navigation}
We adopt and modify the cooperative communication and navigation task
in \cite{lowe2017multi}, where we test \sched~in a simple
one-dimensional grid as in Figure~\ref{fig:ccn}. In CCN, each of the two
agents resides in its one-dimensional grid world. Each agent's goal is
to arrive at a pre-specified destination (denoted by the square with a
star or a heart for Agents 1 and 2, respectively), and they collect a
joint reward when both agents reach their target destination. Each
agent has a zero observation horizon around itself, but it can observe
the situation of the other agent. We introduce heterogeneity into the
scenario, where the agent-destination distance at the beginning of the
task differs across agents.
In our example, Agent 2 is initially located at a farther place from
its destination, as illustrated in Figure~\ref{fig:ccn}. The metric
used to gauge the performance of \sched~is the number of time steps
taken to complete the CCN task.

\subsection{Experiment Details}
Table~\ref{table:hyperparameter} shows the values of the hyperparameters for the CCN and the PP task. 
We use Adam optimizer to update network parameters and soft target update to update target network. The structure of the networks is the same across tasks.
For the critic, we used three hidden layers, and the critic between the scheduler and the action selector shares the first two layers. For the actor, we use one hidden layer; for the encoder and the weight generator, three hidden layers each. Networks use rectified linear units for all hidden layers.
Because the complexity of the two tasks differ, we sized the hidden layers differently. The actor network and the critic network for the CCN have hidden layers
with 8 units and 16 units, respectively. The actor network and the critic network for the PP have hidden layers with 32 units and 64 units, respectively.

\begin{table}[h!]
  \centering
  \caption{\small List of hyperparameters}
  \label{table:hyperparameter}
  \begin{tabular}{@{}lllll@{}}
  \toprule
  \textbf{Hyperparameter}       & \textbf{Value} & \multicolumn{3}{l}{\textbf{Description}}                                \\ \midrule
  training step                 & 750000         & \multicolumn{3}{l}{Maximum time steps until the end of training}        \\
  episode length                & 1000            & \multicolumn{3}{l}{Maximum time steps per episode}                      \\
  discount factor               & 0.9            & \multicolumn{3}{l}{Importance of future rewards}                        \\
  learning rate for actor       & 0.00001        & \multicolumn{3}{l}{Actor network learning rate used by Adam optimizer}  \\
  learning rate for critic      & 0.0001         & \multicolumn{3}{l}{Critic network learning rate used by Adam optimizer} \\
  target update rate            & 0.05           & \multicolumn{3}{l}{Target network update rate to track learned network} \\
  entropy regularization weight & 0.01           & \multicolumn{3}{l}{Weight of regularization to encourage exploration}   \\
  \bottomrule
  \end{tabular}
  \end{table}


\newpage
\section{Additional Experiment Results}
\label{sec:additional-results}

\subsection{Predator and Prey}



\begin{figure}[t!]
\captionsetup[subfigure]{justification=centering}
  \begin{center}
\hspace{-0.2cm}
    \subfloat[\scriptsize Comm. bandwidth $L$ \newline ($k=1$)]{\label{fig:l_var_add}
      \includegraphics[width=0.3\columnwidth]{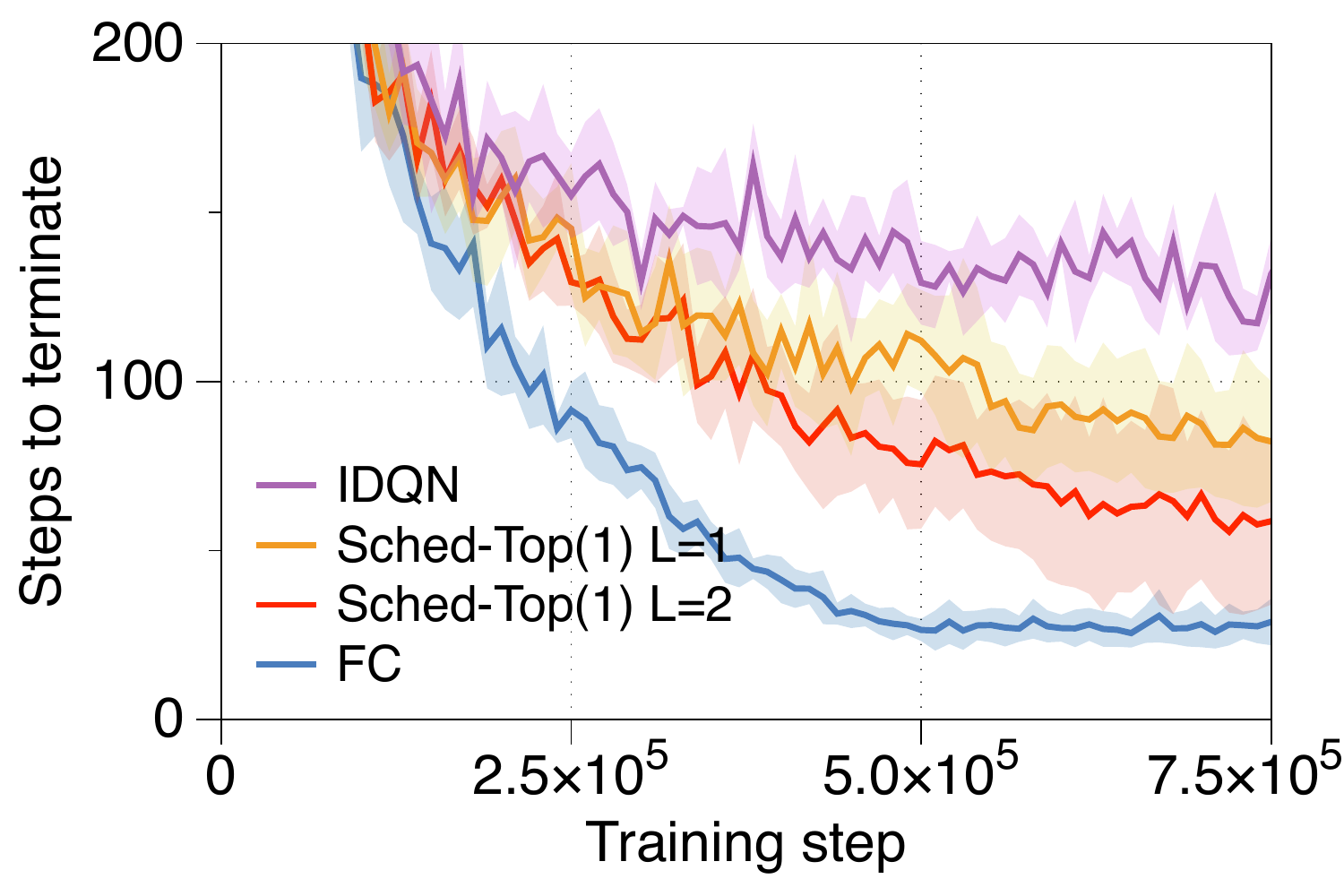} }
    \subfloat[\scriptsize Sched. constraint $k$ \newline ($l=1$)]
    {\label{fig:k_var_add}
      \includegraphics[width=0.3\columnwidth]{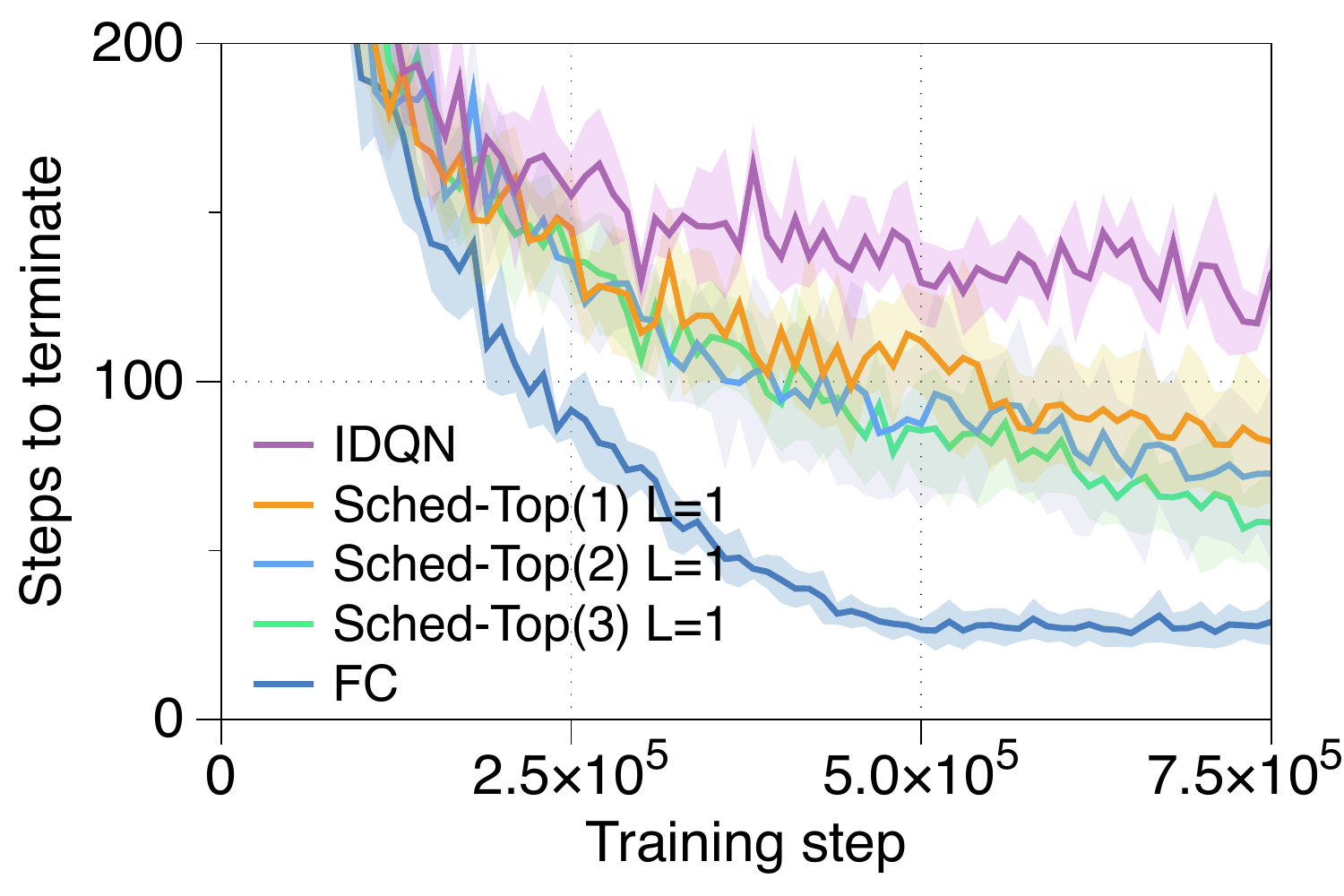} }
   \subfloat[\scriptsize \sched ~vs. AE  \newline ($k=1$ and $l=1$)] {\label{fig:ae_add}
      \includegraphics[width=0.3\columnwidth]{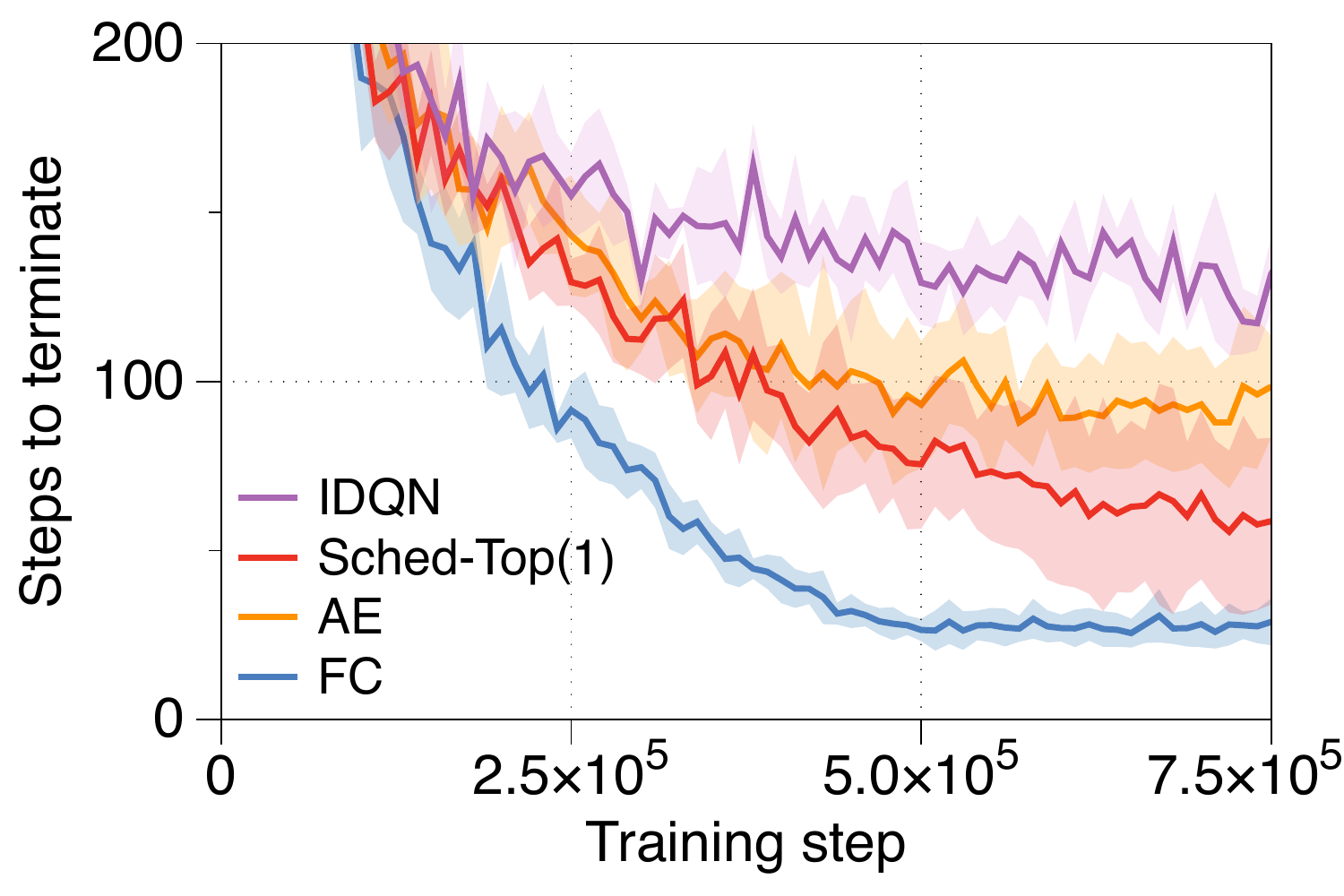} }
\end{center}
\caption{\small Performance evaluation of \sched. The graphs show the
  average time taken to complete the task, where shorter time is
  better for the agents.}
\label{fig:result_add}
\end{figure}

\myparagraph{Impact of bandwidth ($L$) and number of schedulable
  agents ($K$)} Due to communication constraints, only $k$ agents can
communicate and scheduled agents can broadcast their message, each of
which has a limited size $l$ due to bandwidth constraints. We see the
impact of $l$ and $k$ on the performance in
Figure~\ref{fig:l_var_add}. As $L$ increases, more information can be
encoded into the message, which can be used by other agents to take
action. Since the encoder and the actor are trained to maximize the
shared goal of all agents, they can achieve higher performance with
increasing $l$. In Figure~\ref{fig:k_var_add}, we compare the cases
where $k=1,2,3$, and FC in which all agents can access the medium,
with $l=1$. As we can expect, the general tendency is that the
performance grows as $k$ increases.

\begin{table}[h!]
\caption{\small Performance with/without encoder}
\centering {\small
  \begin{tabular}{ccc}\hline
    FC & \begin{tabular}[c]{@{}l@{}}\sched\\ -Top(1)\end{tabular}
            & \begin{tabular}[c]{@{}l@{}}Schedule w/\\
                auto-encoder\end{tabular} \\ \hline\hline
    1       & 2.030   & 3.408     \\ \hline 
\end{tabular}
\label{table:ae}
}
\end{table}

\myparagraph{Impact of joint scheduling and encoding} To study the
effect of jointly coupling scheduling and encoding, we devise a
comparison against a pre-trained auto-encoder \citep{bourlard1988auto,
  hinton1994autoencoders}. An auto-encoder was trained ahead of time,
and the encoder part of this auto-encoder was placed in the Actor's
ENC module in Figure \ref{fig:arch}. The encoder part is not trained
further while training the other parts of network. Henceforth, we name
this modified Actor ``AE''. Figure~\ref{fig:ae_add} shows the learning
curve of AE and other baselines. Table~\ref{table:ae} highlights the
impact of joint scheduling and encoding. The numbers shown are the
performance metric normalized to the FC case in the PP environment.
While \sched -Top(1) took only 2.030 times as long as FC to finish the
PP task, the AE-equipped actor took 3.408 times as long as FC.  This
lets us ascertain that utilizing a pre-trained auto-encoder deprives
the agent of the benefit of joint the scheduler and encoder neural
network in \sched.

\myparagraph{What messages agents broadcast}

\begin{figure}[h!]
  \centering
    \subfloat[Messages for different relative location of pery.]{\label{fig:msg_o}
  \includegraphics[width=0.4\columnwidth]{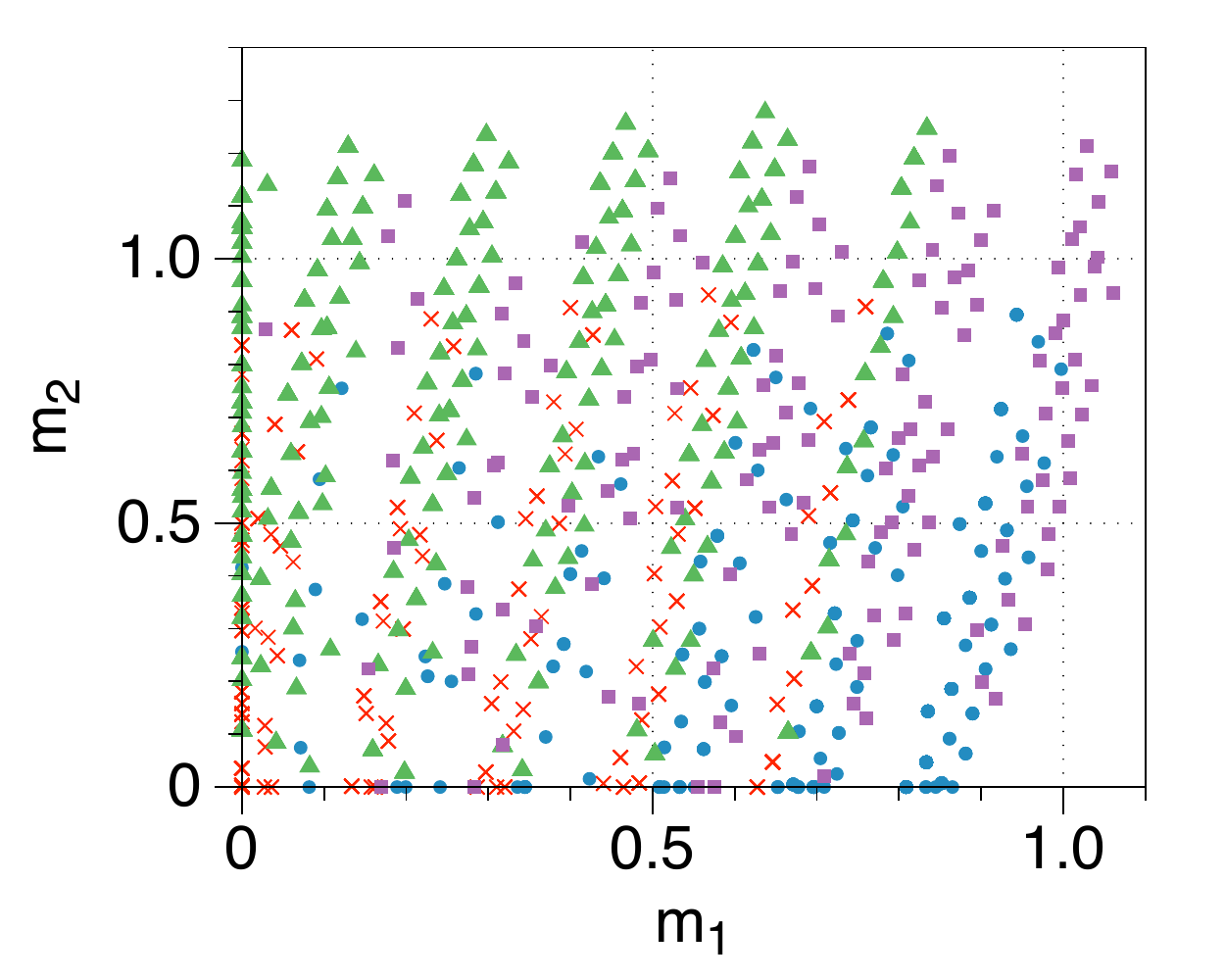}}
\hspace{0.5cm}
    \subfloat[Messages for different  location of agent.]{\label{fig:msg_a}
  \includegraphics[width=0.4\columnwidth]{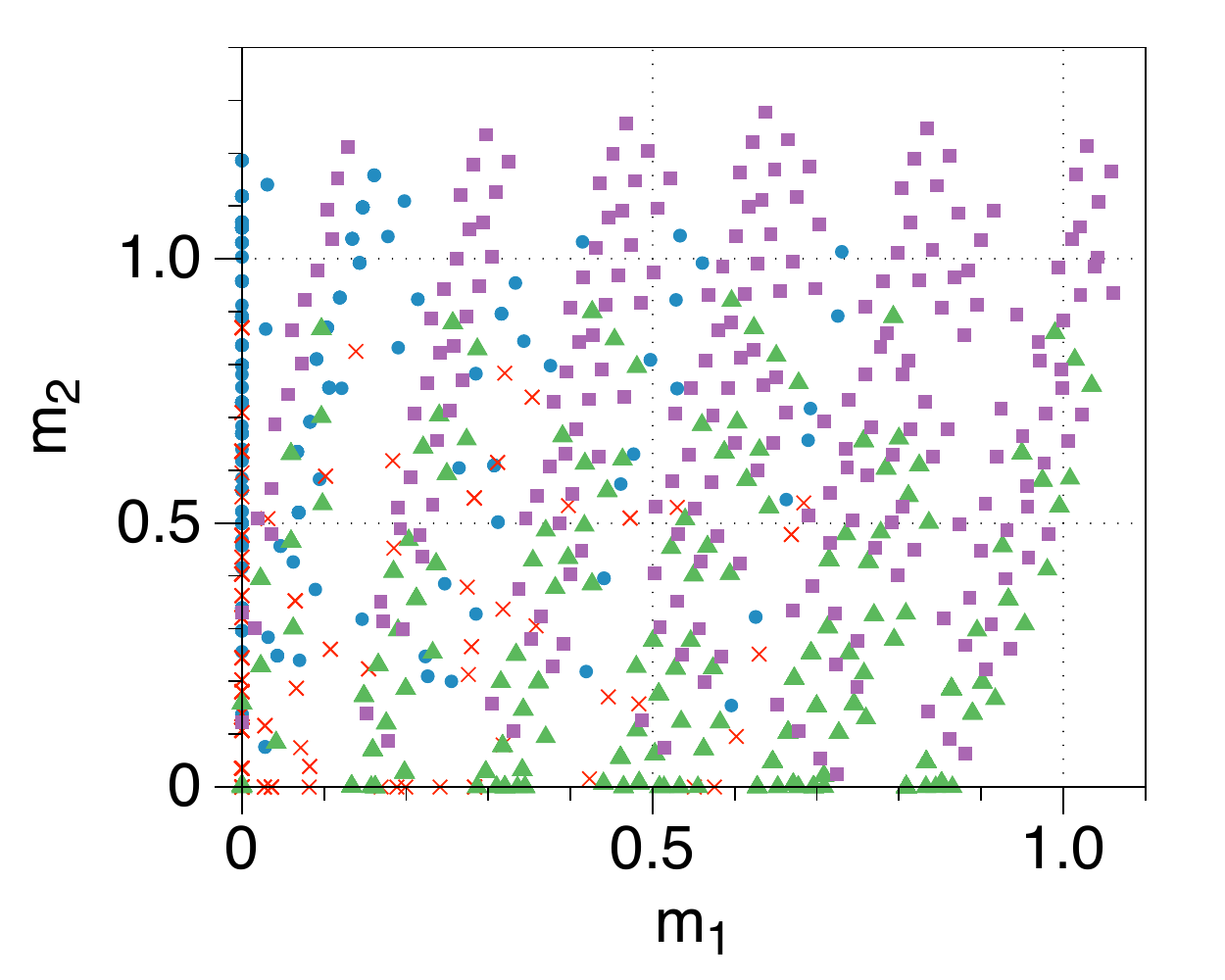}} 
  \vspace{-0.2cm}
\caption{Projection of encoded messages into 2D plane in PP. }
  \vspace{-0.3cm}
\label{fig:msg_add}
\end{figure}

In Section \ref{sec:pp}, we attempted to understand what the predator
agents communicate when performing PP task where $k=1$ and $l=2$. In
this section, we look into the message in
detail. Figure~\ref{fig:msg_add} shows the projections of the messages
generated by the scheduled agent based on its own observation. In the
PP task, the most important information is the location of the prey,
and this can be estimated from the observation of other agents. Thus,
we are interested in the location information of the prey and other
agents. We classify the message into four classes based on which
quadrant the prey and the predator are included, and mark each class
with different colors. Figure~\ref{fig:msg_o} shows the messages for
different relative location of prey for agents' observation, and
Figure~\ref{fig:msg_a} shows the messages for different locations of
the agent who sends the message. We can observe that there is some
general trend in the message according to the class. We thus conclude
that if the agents observe the prey, they encode into the message the
relevant information that is helpful to estimate the location of the
prey.  The agents who receive this message interpret the message to
select action.

\subsection{Partial Observability Issue in \sched}

In MARL, partial observability issue is one of the major problems, and
there are two typical ways to tackle this issue.  First, using RNN
structure to indirectly remember the history can alleviate the partial
observability issues. Another way is to use the observations of other
agents through communication among them. In this paper, we focused
more on the latter because the goal of this paper is to show the
importance of learning to schedule in a practical communication
environment in which the shared medium contention is inevitable.

  \begin{wrapfigure}{R}{0.39\textwidth}
    \centering
  \vspace{-0.4cm}
  \includegraphics[width=0.36\columnwidth]{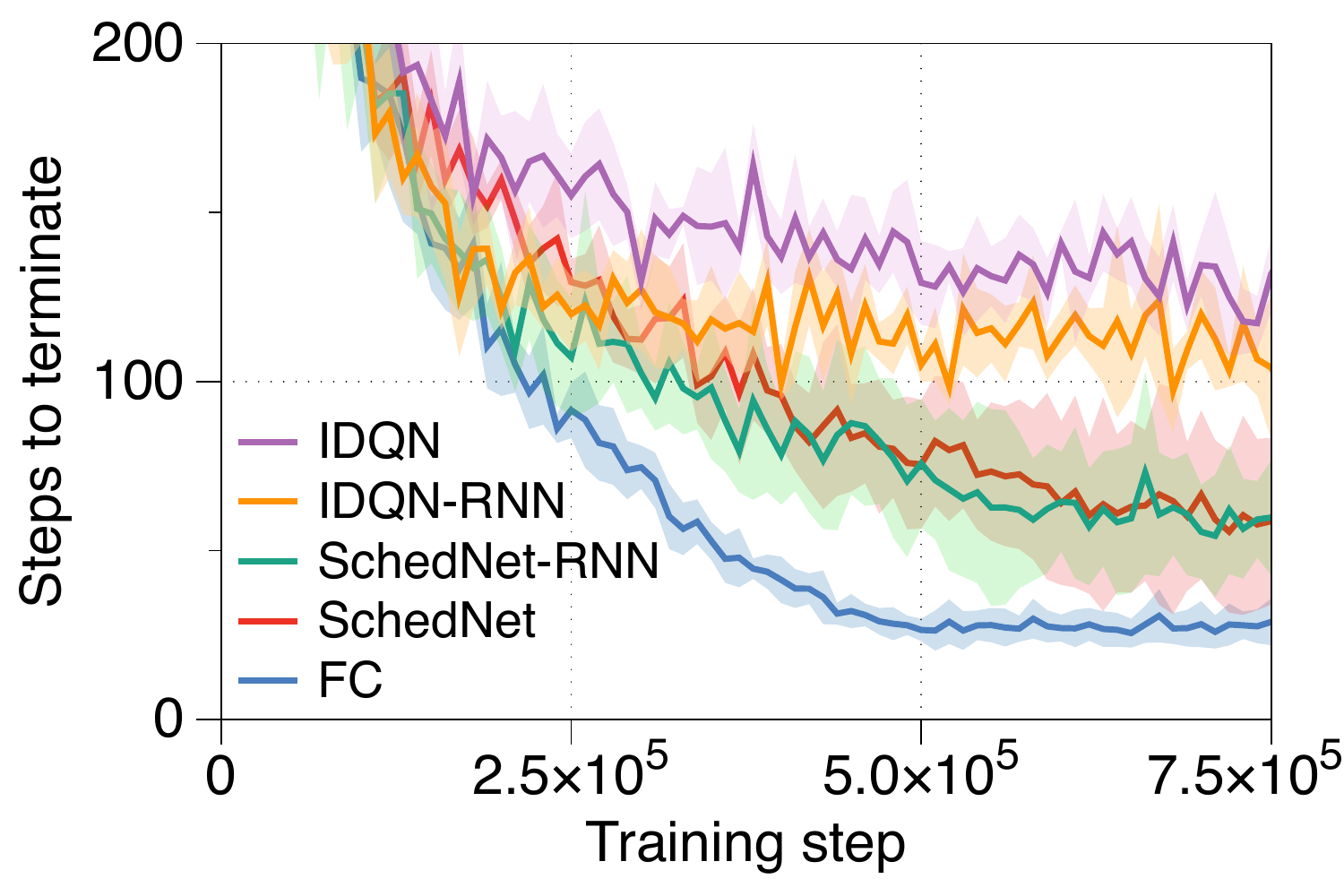}
  \vspace{-0.1cm} \captionof{figure}{\small Impact of applying RNN
    ($k=1$ and $l=2$)}
  \label{fig:rnn}
  \vspace{-0.2cm}
\end{wrapfigure}

Enlarging the observation through communication is somewhat orthogonal
to considering temporal correlation. Thus, we can easily merge
SchedNet with RNN which can be appropriate to some partially
observable environments. We add one GRU layer into each of individual
encoder, action selector, and weight generator of each agent, where
each GRU cell has 64 hidden nodes.

Figure ~\ref{fig:rnn} shows the result of applying RNN. We implement
IDQN with RNN, and the results show that the average steps to complete
tasks of IDQN with RNN is slightly smaller than that of IDQN with
feed-forward network. In this case, RNN helps to improve the
performance by tackling the partial observable issue. On the other
hand, SchedNet-RNN and SchedNet achieve similar performance. We think
that the communication in SchedNet somewhat resolves the partial
observable issues, so the impact of considering temporal correlation
with RNN is relatively small. Although applying RNN to SchedNet is not
really that helpful in this simple environment, we expect that in a
more complex environment, using the recurrent connection is more
helpful.

\subsection{Cooperative Communication and Navigation}
\label{sec:coop-comm-navig}

  \begin{wrapfigure}{R}{0.39\textwidth}
    \centering
  \vspace{-0.4cm}
  \includegraphics[width=0.36\columnwidth]{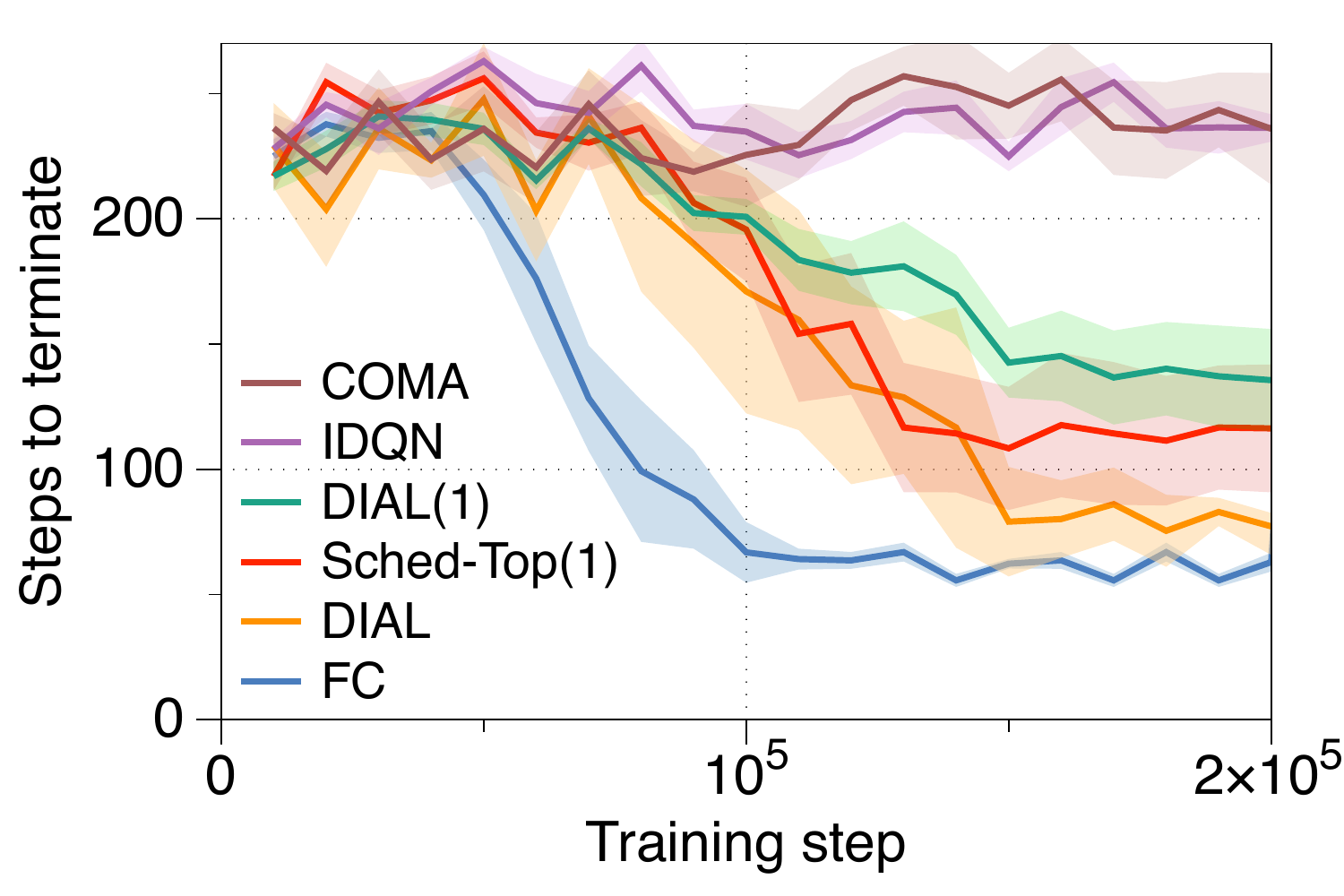}
  \vspace{-0.1cm} \captionof{figure}{\small Comparison with other baselines}
  \label{fig:baseline_ccn}
  \vspace{-0.2cm}
\end{wrapfigure}

\paragraph{Result in CCN} Figure \ref{fig:baseline_ccn} illustrates
the learning curve of 200,000 steps in CCN.  In FC, since all agents
can broadcast their message during execution, they achieve the best
performance. IDQN and COMA in which no communication is allowed, take
a longer time to complete the task compared to other baselines. The
performances of both are similar because no cooperation can be
achieved without the exchange of observations in this environment.  As
expected, \sched~ and DIAL outperform IDQN and COMA.  Although DIAL
works well when there is no contention constraint, under the
contention constraint, the average number of steps to complete the
task in DIAL(1) is larger than that of \sched -Top(1).  This result
shows the same tendency with the result in PP environment.


\section{Scheduler for Distributed Execution}
\label{sec:schedule}

\noindent{\bf Issues.} The role of the scheduler is to consider the
constraint due to accessing a shared medium, so that only $k < n$
agents may broadcast their encoded messages. $k$ is
determined by the wireless communication environment. For example,
under a single wireless channel environment where each agent is
located in other agents' interference range, $k=1$.  Although the
number of agents that can be simultaneously scheduled is somewhat more
complex, we abstract it with a single number $k$ because the goal of
this paper lies in studying the importance of considering scheduling
constraints.

There are two key challenges in designing the scheduler: {\em (i)} how to
schedule agents in a distributed manner for decentralized execution, and
{\em (ii)} how to strike a good balance between simplicity in implementation
and training, and the integrity of reflecting the current practice of
MAC (Medium Access Control) protocols.

\begin{figure}[h!]
  \centering
  \includegraphics[width=0.5\columnwidth]{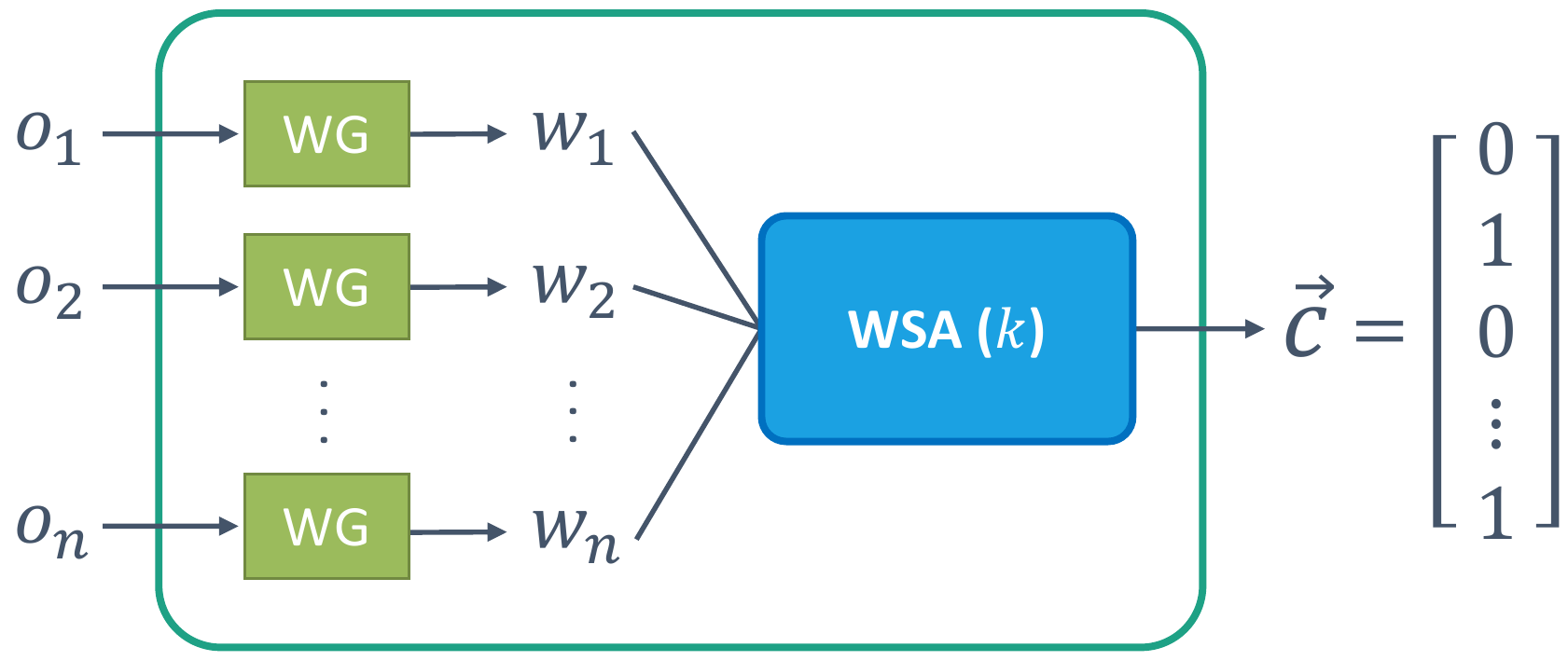}
  \vspace{-0.2cm}
\caption{\small Proposed scheduling architecture. 
  Each agent $i$ calculates its scheduling weight $w_i$
  from weight generator (WG), and
  the corresponding scheduling profile ${\bm c} \in \{0,1\}^n$
  is determined by the scheduling algorithm (k) (WSA(k)), satisfying the
  condition $||{\bm c}||_1 = k$.}

\label{fig:schedule_add}
\end{figure}

\paragraph{Weight-based scheduling}
To tackle the challenges addressed in the previous paragraph, we
propose a scheduler, called weight-based scheduler (WSA), that works based on each agent's individual weight
coming from its observation. As
shown in Figure~\ref{fig:schedule_add}, the role of WSA is to map from
$\vec{w}=[w_i]_n$ to $\vec{c}.$ This scheduling is extremely simple,
but more importantly, highly amenable to the philosophy of distributed
execution. The remaining checkpoint is whether this principle is
capable of efficiently approximating practical wireless scheduling
protocols. To this end, we consider the following two weight-based
scheduling algorithms among many different protocols that could be
devised:
\begin{compactitem}[$\circ$]

\item {\em Top($k$).} Selecting top $k$ agents in terms of their weight values. 

\item {\em Softmax($k$).} Computing softmax values $\sigma({\bm w})_i = \frac{e^{w_i}}{\sum_{j=1}^{n}e^{w_j}}$ for each agent $i,$ and then randomly selecting  $k$ agents with probability in proportion to their softmax values. 
\end{compactitem}

Top($k$) can be a nice abstraction of
the MaxWeight~\citep{tassiulas1992stability} scheduling principle or its
distributed approximation \citep{yy_scheduling}, in which case it is
known that different choices of weight values result in achieving
different performance metrics, {\em e.g.}, using the amount of messages
queued for being transmitted as weight. Softmax($k$) can be a
simplified model of CSMA (Carrier Sense Multiple Access), which forms
a basis of 802.11 Wi-Fi.  Due to space limitation, we refer the reader
to \citet{jiang2010distributed} for detail. We now present how {\em
  Top($k$)} and {\em Softmax($k$)} work.

\subsection{Carrier Sense Multiple Access (CSMA)}
CSMA is the one of typical distributed MAC scheduling in wireless
communication system. To show the feasibility of scheduling {\em
  Top($k$)} and {\em Softmax($k$)} in a distributed manner, we will
explain the variant of CSMA. In this section, we first present the
concept of CSMA.

\myparagraph{How does CSMA work?} The key idea of CSMA is ``listen
before transmit''. Under a CSMA algorithm, prior to trying to transmit
a packet, senders first check whether the medium is busy or idle, and
then transmit the packet only when the medium is sensed as idle, {\em i.e.}, no
one is using the channel. To control the aggressiveness of such medium
access, each sender maintains a backoff timer, which is set to a
certain value based on a pre-defined rule. The timer runs only when
the medium is idle, and stops otherwise. With the backoff timer, links
try to avoid collisions by the following procedure:
\begin{itemize}
\item Each sender does not start transmission immediately when the
  medium is sensed idle, but keeps silent until its backoff timer
  expires.
\item After a sender grabs the channel, it holds the channel for some
  duration, called the holding time.
\end{itemize}
Depending on how to choose the backoff and holding times, there can be
many variants of CSMA that work for various purposes such as fairness
and throughput. Two examples of these, {\em Top($k$)}
and {\em Softmax($k$)}, are introduced in the following sections.

\subsection{A version of {\em Distributed Top($k$)} }
In this subsection, we introduce a simple distributed scheduling
algorithm, called {\em Distributed Top($k$)}, which can work with
\sched -Top($k$). It is based on CSMA where each sender determines
backoff and holding times as follows.  In \sched, each agent generates
the scheduling weight $w$ based on its own observation. The agent sets its
backoff time as $1-w$ where $w$ is its schedule weight, and it waits
for backoff time before it tries to broadcast its message. Once it
successfully broadcasts the message, it immediately releases the channel.
Thus, the agent with the highest $w$ can grab the channel in a
decentralized manner without any message passing. By repeating this
for $k$ times, we can realize decentralized Top($k$) scheduling.

To show the feasibility of distributed scheduling, we implemented the
Distributed Top($k$) on Contiki network simulator~\citep{contiki} and
run the trained agents for the PP task. In our experiment, Top($k$) agents
are successfully scheduled 98\% of the time, and the 2\% failures are
due to probabilistic collisions in which one of the colliding agents is
randomly scheduled by the default collision avoidance mechanism implemented in
Contiki. In this case, agents achieve 98.9\% performance compared to
the case where Top($k$) agents are ideally scheduled.

\subsection{oCSMA Algorithm and {\em Softmax($k$)}}
In this section, we explain the relation between {\em Softmax($k$)} and
the existing CSMA-based wireless MAC protocols, called oCSMA. When we
use {\em Softmax($k$)} in the case of $k = 1$, the scheduling
algorithm directly relates to the channel selection probability of
oCSMA algorithms. First, we explain how it works and show that the
resulting channel access probability has a same form with {\em
  Softmax($k$)}.

\myparagraph{How does oCSMA work?} It is also based on the basic CSMA
algorithm. Once each agent generates its scheduling weight $w_i$, it sets
$b_i$ and $h_i$ to satisfy $w_i = \log(b_i h_i)$. It sets its backoff
and holding times following exponential distributions with means
$1/b_i$ and $h_i$, respectively. Based on these backoff and holding
times, each agent runs the oCSMA algorithm. In this case, if all agents are
in the communication range, the probability that agent $i$ is
scheduled over time is as follows:
$$s_i (\bm{w}) 
= \frac{\exp(w_i)}{\sum_{j=1}^{n} \exp(w_j)}.$$
 We refer the readers to \citet{jang} for detail.

\end{document}